# A Comprehensive Scoping Review of Bayesian Networks in Healthcare: Past, Present and Future

Evangelia Kyrimi †

School of Electronic Engineering and Computer Science (EECS), Queen Mary University of London, London, United Kingdom, e.kyrimi@qmul.ac.uk

Scott McLachlan

School of Electronic Engineering and Computer Science (EECS), Queen Mary University of London, London, United Kingdom, s.mclachlan@qmul.ac.uk
Health informatics and Knowledge Engineering Research (HiKER) Group

Kudakwashe Dube

School of Fundamental Sciences, Massey University, Palmerston North, New Zealand, k.dube@massey.ac.nz
Health informatics and Knowledge Engineering Research (HiKER) Group

Mariana R. Neves

School of Electronic Engineering and Computer Science (EECS), Queen Mary University of London, London, United Kingdom, m.r.neves@qmul.ac.uk

Ali Fahmi

School of Electronic Engineering and Computer Science (EECS), Queen Mary University of London, London, United Kingdom, a.fahmi@qmul.ac.uk

Norman Fenton

School of Electronic Engineering and Computer Science (EECS), Queen Mary University of London, London, United Kingdom, n.fenton@qmul.ac.uk

† Corresponding author

## ABSTRACT

No comprehensive review of Bayesian networks (BNs) in healthcare has been published *in the past*, making it difficult to organize the research contributions i*n the present* and identify challenges and neglected areas that need to be addressed *in the future*. This unique and novel scoping review of BNs in healthcare provides an analytical framework for comprehensively characterizing the domain and its current state. The review shows that: (1) BNs in healthcare are not used to their full potential; (2) a generic BN development process is lacking; (3) limitations exist in the way BNs in healthcare are presented in the literature, which impacts understanding, consensus towards systematic methodologies, practice and adoption of BNs; and (4) a gap exists between having an accurate BN and a useful BN that impacts clinical practice. This review empowers researchers and clinicians with an analytical framework and findings that will enable understanding of the need to address the problems of restricted aims of BNs, ad hoc BN development methods, and the lack of BN adoption in practice. To map the way forward, the paper proposes future research directions and makes recommendations regarding BN development methods and adoption in practice.

## CCS CONCEPTS

• Mathematics of computing • Probability and statistics • Probabilistic inference problems • Bayesian computation

**KEYWORDS**
Bayesian networks, Healthcare, Clinical decision support, Scoping review, Survey

# 1 Introduction

Bayesian Networks (BNs), also described as *causal probabilistic models* or *belief networks*, are directed acyclic graphical (DAG) models (Pearl 1988). Early attempts to use Bayesian analysis for medical problems were generally considered unsuccessful due to the necessary Bayesian inference being computationally intractable (Cooper 1990). Development of efficient BN inference propagation algorithms (Pearl 1988), (Heckerman 1995) and advances in computational power since, have made it possible to develop BNs capable of addressing real-world decision support problems. This led to renewed research interest which identified many thousands of publications describing BN solutions in the context of healthcare.

The immense volume of published material demonstrates an enormous and rapidly increasing appetite for BNs as a reasoning tool to support healthcare delivery. With such significant research interest, it is important to properly organize and summarize the range of research contributions and enable identification of research gaps and future research directions. However, apart from a small number of micro-reviews about BNs for specific medical conditions (Bielza and Larranaga 2014) and one epidemiology-focused review on DAGs that is yet to be peer reviewed (Tennant et al. 2019), the domain lacks any form of systematic or scoping review. Additionally, evidence for their adoption in practice remains extremely limited. There is a significant gap between developing a claimed accurate model and demonstrating its clinical usefulness and actual impact on healthcare decision-making (Wyatt and Altman 1995; Blackmore 2005; Reilly and Evans 2006; Moons et al. 2009; Adams and Leveson 2012).

This work presents a scoping review that examines and evaluates BNs in healthcare with specific regard to identifying the: (1) targeted *decision support*, (2*) modelling approach*, and (3) *usefulness* in practice. We believe that combined, these provide a global assessment of the research domain. To the best of our knowledge, this is the first scoping review on BNs in healthcare.

The rest of this paper is organized as follows: Section 2 presents necessary background material related to BNs. Section 3 provides the motivation for undertaking this scoping review. The methodology followed and the obtained results are described in Sections 4 and 5, respectively. A list with all the major findings and a detailed discussion on the findings as well as the strength and limitation of this review are described in Sections 6 and 7, respectively. Section 8 offers a list of recommendations for the authors, while Section 9 shows future research directions. Finally, a summary is provided in Section 10.

# 2 Background

In order to ground our objectives for this review it is necessary that we review and describe background material regarding: (1) the *fundamentals* of BNs, (2) the *reasoning process* in BNs, and (3) the *significance* of BNs. Readers familiar with some or all of the described preliminaries may select to skip these subsections.

## 2.1 Bayesian fundamentals

In late 1750, the English Mathematician and Reverend Thomas Bayes developed the now-famous Bayes' theorem (Bayes, 1763). His study focused on how prior belief can be updated based on new evidence; a process now known as Bayesian inference, or Bayesian reasoning. Initial belief in the likelihood of a particular event or outcome are described as *prior probabilities* (for a hypothesis H we write this as P(H)); while updated beliefs in the light of new evidence E are *posterior probabilities* (written P(H|E) representing the conditional probability of H given E). This initial and updated belief process is common to everyday life. For example, when clinicians perform differential diagnosis in circumstances where multiple diseases present with similar symptoms. As new evidence from diagnostic tests and clinical examination of the patient become available, clinicians update their belief on what they consider to be the illness causing the patient's poor health.

Bayes' theorem is a simple equation for calculating the posterior probability of a hypothesis H in terms of the prior probability, and the probability of the evidence conditional on H and its negation not H:

$$P(H|E) = \frac{P(E|H)P(H)}{P(E|H)P(H) + P(E|not\ H)P(not\ H)}$$

BNs are based on Bayes' theorem and use a graphical approach for *compact representation of multivariate probability distributions* and *efficient reasoning under uncertainty* (Pearl, 1993). At its basic level, a BN is a directed acyclic graph with qualitative and quantitative parts (Pearl, 1993). The qualitative part is the BN *structure* comprised of nodes representing random variables (discrete or continuous) and directed arcs representing causal or influential relationships. If a directed arc connects variables *A* and *B*, such as *A*→*B*, then *A* is called parent node or ancestor of *B* and *B* is a child node or a successor of *A*. The qualitative part is the BN *parameters* comprised of a set of conditional probability functions associated with each node – captured by a Node Probability Table (NPT) - to represent the conditional probability distribution of each node in the BN given its parents.

Both BN structure and parameters can be prepared using: (a) automated learning from data if sufficient data are available (Dranca et al. 2018); (b) a manual "by-hand" approach using knowledge elicitation methods to capture domain expert knowledge and extracting the necessary information from literature (Philipp, Beyerer, and Fischer 2017); or, (c) through a combination of both (Flores et al. 2011).

## 2.2 Reasoning process

As values for all NPTs are provided the BN becomes fully parameterized (Figure 1a) such that it can be used in a variety of probabilistic reasoning processes (through applying Bayes theorem). There are three ways of reasoning with BNs: (i) *observational reasoning* that is reasoning from evidence; (ii) *interventional reasoning*; and (iii) *counterfactual reasoning*.

*Reasoning from evidence* describes a joint distribution over possible observed events (Woodward 2003). Evidence represents the real world, what happened. Once evidence is observed into the BN, the probabilities of remaining unobserved variables can be updated. There are two approaches to reasoning from evidence:

1. *Causal reasoning*: where the reasoning process follows the direction of the arc. For example: knowing that the patient is a smoker increases the probability of lung cancer from 6.4% (Figure 1a) to 10% (Figure 1b), which in turn increases the probability of a positive X-ray result from 15.4% to 18.5%.
2. *Diagnostic reasoning*: where the reasoning process is counter to the direction of the arc. For example: knowing that the patient's X-ray is positive increases the probability of lung cancer (Figure 1c).

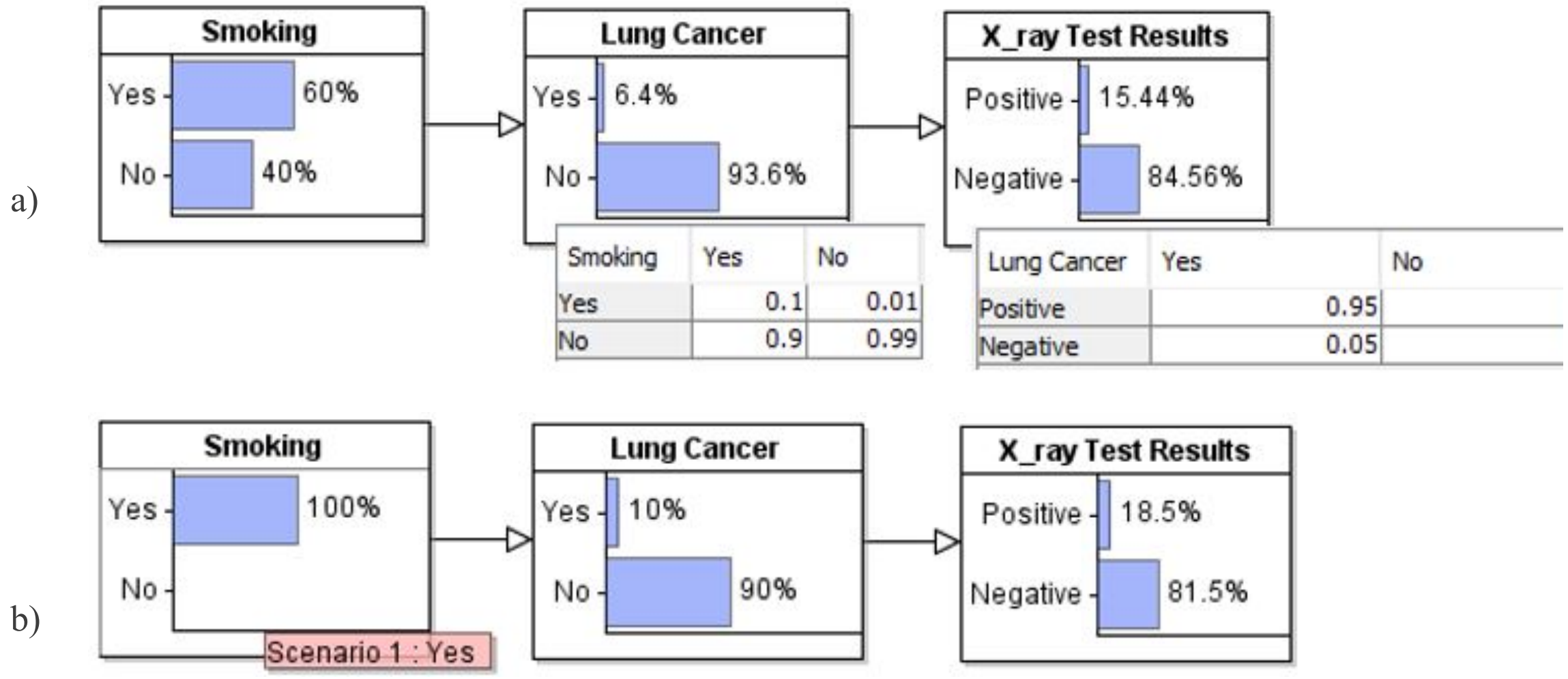

c)

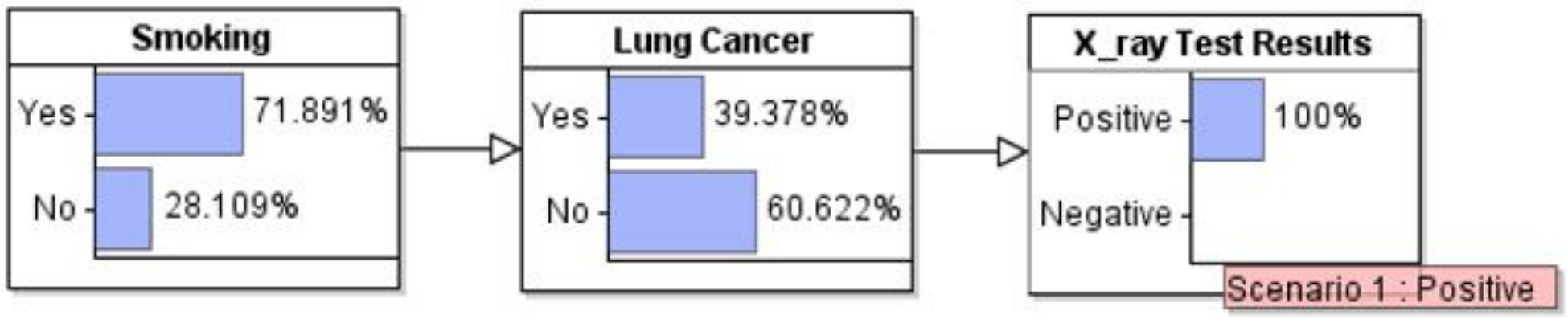

**Figure 1: (a) A three-node BN example with Node Probability Tables shown, (b) A BN example, where causal reasoning is performed, (c) A BN example, where diagnostic reasoning is performed.**

BNs can also be used to answer hypothetical questions such as *what will happen if an intervention is made*. *Interventional reasoning* can only be performed when relationships among the variables are causal because an intervention is an exogenous action that fixes the state of the variable we have intervened upon, thus, making it independent of its causes (Pearl 1993; Pearl 2009; Shpitser and Pearl 2008). Contrary to reasoning about evidence, *interventional reasoning* does not allow diagnostic reasoning from the intervened variable (Hagmayer et al. 2007). For instance, when we observe a high body temperature on the thermometer, we can argue that we have a fever. However, if we arbitrarily start rubbing the thermometer to reach a specific temperature, we can no longer argue that we have a fever. According to Pearl (Pearl 2009), an externally imposed intervention is presented using the *do operator*. The process of making the intervened variable independent of its causes by removing all the edges pointing towards that variable is known as *graph surgery*. BNs can also be used to answer *counterfactual questions* such as *what would have happened if events other than those observed had occured*. Considering contrary-to-facts scenarios means imagining alternatives to reality. Although sometimes claimed otherwise (Dawid 2000), counterfactual reasoning is a natural reasoning process and is believed to be an ability unique to the human mind (Sloman 2013). In BNs, counterfactual reasoning combines both evidence and interventions. Using the process of *twin-networks* proposed by Pearl (Pearl 2009), the actual world is modelled based on evidence while the hypothetical counterfactual world is altered using interventions. In a twin network both networks have identical structures except for the arrows towards the variable that we intervene upon, which are missing in the hypothetical world. The posterior probabilities of the variables, which remain the same in both worlds, are called *background variables,* and are shared between the two networks. The main difference between counterfactuals and interventions is that for the former we know the values that some or all the variables had in the actual world. By contrast, when we intervene, we are unaware of the values of the predecessor variables in the network.

## 2.3 Significance in healthcare

Clinical decision making is a complex evolving process where evidence is gathered and diagnostic and treatment decisions are made (Tiffen, Corbridge, and Slimmer 2014). Through consideration of signs, symptoms, a patient's history and any test results, clinicians try to resolve two main questions: 'What is the problem?' and 'How can we solve it?'. Symptoms, diagnostic tests, diseases, treatment options, our complex human physiology and increased uncertainty can make clinical decision-making challenging. While clinicians can be good decision-makers, it can be difficult to combine all available evidence and accurately reason under conditions of such uncertainty (Bornstein and Emler 2001).

Many clinical decision support systems (CDSS) have been developed to support clinicians in their decision-making role (Druzdzel and Flynn 2002; Beattie and Nelson 2006; Patel et al. 2009; Adams and Leveson 2012; Shortliffe and Cimino 2013). These can range from simple scoring systems to complicated multivariate regression models, neural networks, decision trees and graphical probabilistic models like BNs. BNs are being recognized as powerful tools for risk analysis and decision support for real-world problems and have become a popular CDSS model for healthcare application (Lucas 2001; Lucas, van der Gaag, and Abu-Hanna 2004). Their popularity in healthcare results from their ability to: (i) model complex problems with causal dependencies where a significant degree of uncertainty is involved; (ii) combine different sources of information such as data and experts' judgement; (iii) have an interpretable graphical structure; and, (iv) model interventions and reason both diagnostically and prognostically.

# 3 Motivation

The following conclusions were formulated from our preliminary review of BNs in healthcare (Kyrimi et al. 2020):

1. The body of literature on BNs in healthcare is large and rapidly increasing, indicating a significant research interest.
2. The high volume of published works and lack of attempts to systematically review the domain are significant motivating factors for undertaking a systematic or scoping review to summarize the literature and highlight research gaps and future directions.
3. The process of developing a BN is rarely described clearly in the literature to the point of being repeatable.
4. Despite the obvious and widely accepted benefits of BNs and the enormous number of publications, there has been negligible clinical adoption of the BNs described in published articles.

These conclusions provide a strong case to support conducting a larger scale review to identify causes for some of the highlighted issues and the proposal of a way forward towards their resolution. Indeed, the lack of such systematic or scoping reviews is likely to have led to duplication of effort and repetition of methodological mistakes, while at the same time allowed important research gaps to remain hidden.

# 4 Methodology

## 4.1 Literature search

A comprehensive search of the major health and health informatics literature databases including PubMed, MedLine, ScienceDirect, Scopus, DOAJ and Elsevier was performed using a selection of keywords arranged in the following general search query:

"(((Bayes OR Bayesian) AND network) OR (probabilistic AND graphical AND model)) AND (medical OR clinical)"

Terms such as *Bayesian networks* or *graphical probabilistic models* were used here because they are widely observed in the targeted literature. Different ways for explaining the medical condition do occur, in that in some papers the exact condition is mentioned while in others broader terms such as *medical* or *clinical application*, *medical* or *clinical condition*, or *medical* or *clinical setting* are used. Our scoping review settled on the broader terms *medical* or *clinical* as they were found in a wider collection of papers. Searching for specific medical conditions would have been impractical as there are many thousands of distinct known conditions.

Due to the high number of articles returned, further scrutiny was applied to narrow the collection to the most relevant articles for this study. This was achieved by assessing whether the described keywords were present in the abstract. Additional screening was conducted to exclude papers published outside the period 2012-2018, those that were not published in English, or those whose primary content was not healthcare related. The remaining eligible papers were those possessing all the following identified characteristics. They:

1. Describe a genuine BN model or BN modeling process or BN adoption process.
2. Are targeted clearly at a medical condition or application.
3. Are intended to support clinicians or patients in decision making or prediction

## 4.2 Analytical framework for characterizing BNs in healthcare

Figure 2 presents the analytical framework that has been developed in this work for use in characterizing the domain of BNs in healthcare. This framework also guides the review and analysis of the body of literature in this study. In alignment with this framework, our review plan identified six primary objectives shown in Figure 2 that cover important aspects critical to harnessing BNs in healthcare. The focus of the results reported in this paper are limited to the following three of these objectives (leaving the remaining objectives to more focused investigations that will be presented in later papers):

1. *Objective 1 – Decision support*: This objective involves investigating published BN models focusing on the identification of: (1) the *reasoning processes* which have been defined in Section 2; and (2) the *type of decision support* associated with the reasoning process, which influences and is influenced by the *temporal aspect* of the medical problem.
2. *Objective 2 – Modelling approach*: BN development was identified and broken into two parts, each determined by: (1) the BN structure (variables and arcs), and (2) the conditional probabilities for each variable (parameters). The source for variables, arcs and parameters was identified and recorded. Additionally, we explored whether the paper provided a repeatable overall BN development process along with a clear description of the structure and parameter elicitation or learning process. Finally, the distribution of BN software tools used by authors, where identified, was captured.
3. *Objective 3 – Model usefulness*: The potential usefulness of published BNs to clinical practice was examined. Many researchers have investigated the gap between developing an accurate model and having a useful model that has or can have an actual impact on the clinical decision-making process (Reilly and Evans 2006; Moons et al. 2009; Altman et al. 2009; Adams and Leveson 2012). Based on existing literature (Wyatt and Altman 1995; Blackmore 2005; Toll et al. 2008), the properties presented in Table 1 are considered necessary and should be present to ensure a useful model.

**Table 1: Review framework for model usefulness.**

| Properties | Description |
|---|---|
| *Benefit* | The BN addresses a clinical question of enough importance that has a margin of improvement to justify time by practitioners |
| *Credibility* | The BN has a clear logic and appears to describe what it intends to describe |
| *Accuracy* | The BN has an acceptable predictive performance (internal validation) |
| *Generalizability* | The BN has an acceptable accuracy in datasets other than the original training dataset (temporal and/or external validation) |
| *Usability* | There is an interface adjusted to the user's needs |
| *Impact* | The model must have potential to change patient care |

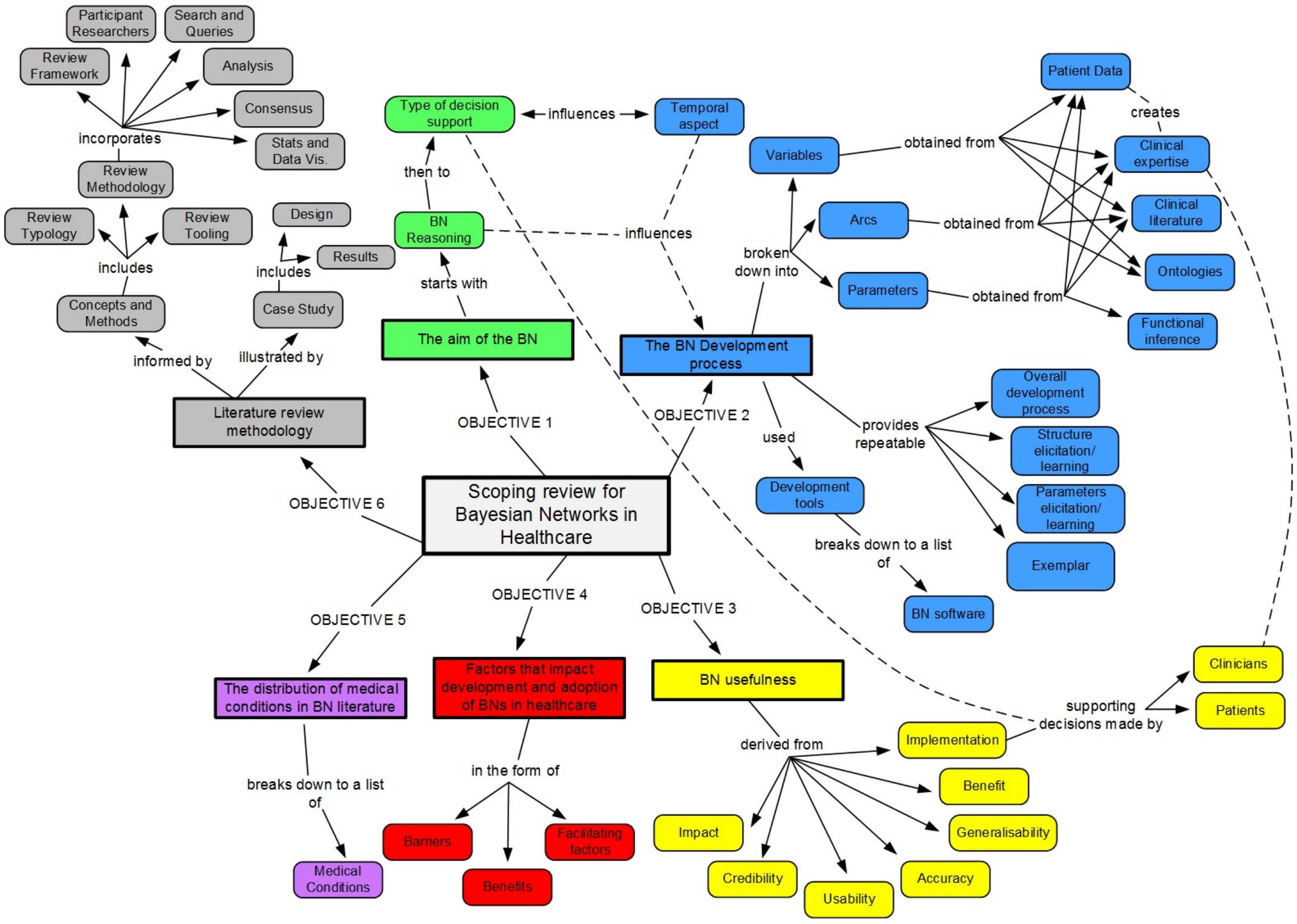

**Figure 2: Framework for characterizing the domain of Bayesian networks in healthcare.**

Each objective identified in Figure 2 captures a unique attribute from the literature on BNs in healthcare. The first three objectives when combined provide a global assessment of the BNs in healthcare domain. As shown in Figure 2, each objective has been broken into a set of attributes which we can observe and record from the literature. These attributes were developed with the input of two decision scientists and refined inductively during a preliminary review conducted by the first two authors (EK and SM) and reported in (Kyrimi et al. 2020). In addition, two decision scientists experienced in developing BNs and who were not involved in the preliminary study were asked to further refine the selected attributes.

The proposed framework was also used to ensure a consistent review of the literature. To further empower a consistent review process, a set of tools that included a systematically designed online literature review questionnaire and a manual for researchers on how to interpret and respond to the review questions was distributed. The manual provided a detailed definition for each attribute and a clear explanation of how to identify that attribute in the literature being reviewed. Additional training sessions were conducted during which 3 papers were reviewed by our reviewers, followed by a thorough discussion on the given answers and how they were derived. Each paper in this review was examined by two reviewers who entered their responses into the secure online survey tool. In cases where responses differed, two senior reviewers (EK and SM) reviewed the paper collaboratively and resolved the conflict by consensus.

# 5 Results

## 5.1 Literature search and collection results

Our initial search identified a collection of 3810 papers for screening. After applying the screening process described in Section 4.1, 462 papers remained for eligibility assessment. As the objective of this review was to investigate: (1) the targeted decision support, (2) the BN modeling process, and (3) the BN usefulness in healthcare, we excluded as ineligible papers whose focus was on any of the following:

1. Bayesian statistics, as opposed to Bayesian networks
2. Simple naive Bayesian or neural networks
3. Network meta-analysis
4. The predictive performance of multiple machine learning algorithms

Those works that did not focus on a medical application were also not eligible for inclusion. 123 papers met the selection and eligibility criteria and remained for inclusion in this review. The results of the literature selection process are presented in diagrammatic form in Figure 3. The full list of these 123 papers is provided in https://pambayesian.org/interim-data/

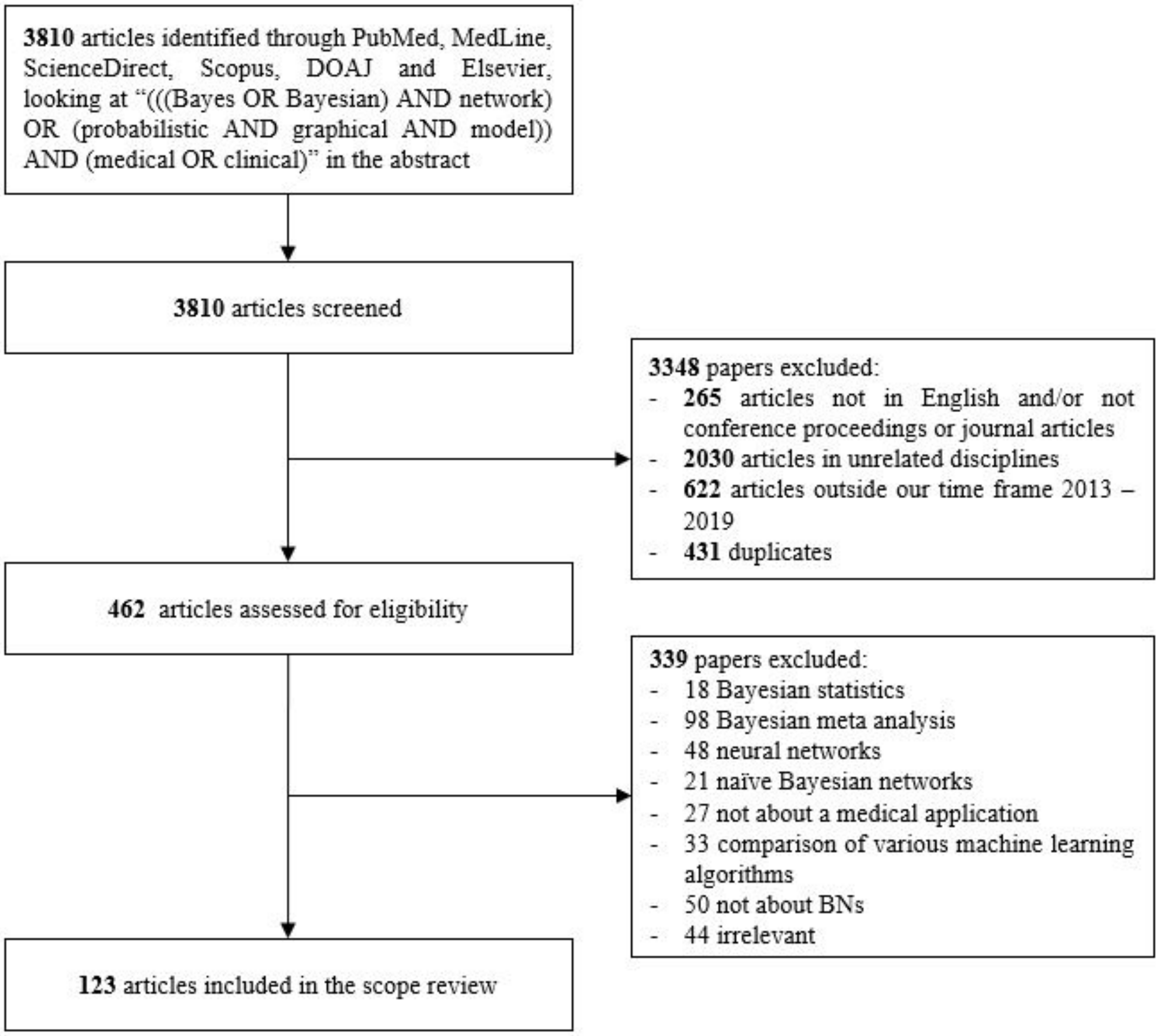

**Figure 3: Scoping review literature selection.**

## 5.2 Literature search and collection results

This section presents results from analysis of included literature with respect to the elements identified for each of the first three objectives of the review as illustrated in Figure 2.

### 5.2.1 Decision support

As shown in Figure 4, the first objective of this review was to identify the aim of medical BN models described in the literature. This section applies the methodology described in Section 4.2 and presents the results of that investigation.

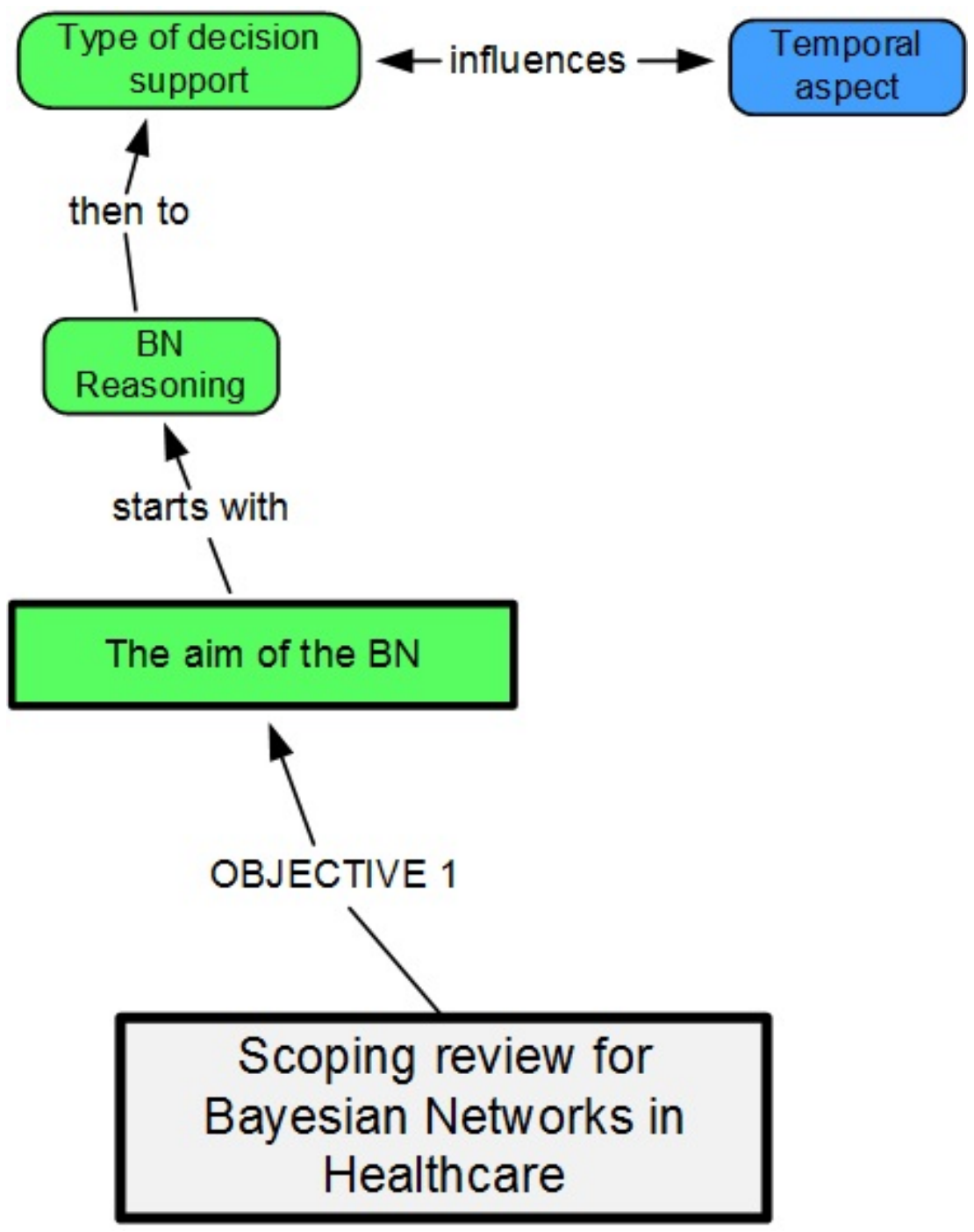

**Figure 4: Concept map for the first objective of the scoping review.**

In the vast majority of papers *observational reasoning* was performed, for example: Gianaroli et al. (2013), Qian, Wang, and Jiang (2014), Wang et al. (2015), Agarwal, Verma, and Mallik (2016), Ong et al. (2017) and Dranca et al. (2018). Figure 5 shows that there is a range of decisions for which observational reasoning may be applied. Diagnosis of a medical condition given a set of known risk factors and observed signs and symptoms is the most frequent, as may be seen in examples such as the works of Karaca et al. (2013), Qian, Wang and Jiang (2014), Khademi and Nedialkov (2015), Refai, Merouani, and Aouras (2016), Kuang et al. (2017), and Liu et al. (2018). For a review and detailed description of the scope of targeted *medical conditions* identified from our literature collection, we refer the reader to (McLachlan et al. 2020). *Interventional reasoning* was much less frequent and was mainly used for treatment selection, such as seen in Sesen et al. (2013), (2014), Liu et al. (2015), Constantinou et al. (2015), Zarikas, Papageorgiou, and Regner (2015). In three papers, namely, the two included works of Constantinou et al. (2015), (2016) and single work from Xu et al. (2018), interventional reasoning was performed for managing acceptable risk through consideration of a number of relevant interventions. In one paper by Neapolitan et al. (2016), interventional reasoning was used to predict prognosis: that is, the likelihood of treatment success. In the papers published by Hernadez-Leal et al. (2013), Cai et al. (2015), Khademi and Nedialkov (2015), Fuster-Parra et al. (2016), Solomon et al. (2016) and Chao et al. (2017), interventional reasoning was mentioned as a potential future use for their developed BNs. While the use of the published BNs to answer *counterfactual questions* was discussed by Solomon et al. (2016) as a potential future direction, it is only in the work of Constantinuou et al. (2015) that the proposed BN was actually applied in a counterfactual setting. Finally, in a significant proportion of the literature, particularly in the works of Hernadez-Leal et al. (2013), Kim et al. (2014), Frohlich et al. (2015), Jin et al. (2016), Zeng, Jiang, and Neapolitan (2016), Lee and Jiang (2017) and Noyes et al. (2018), no true reasoning process was performed. Rather, the BN was used to learn associations or causal relationships from data in order to understand and provide causal attribution to a disease, which relates more to BN model development rather than BN model use or its potential for application in clinical practice.

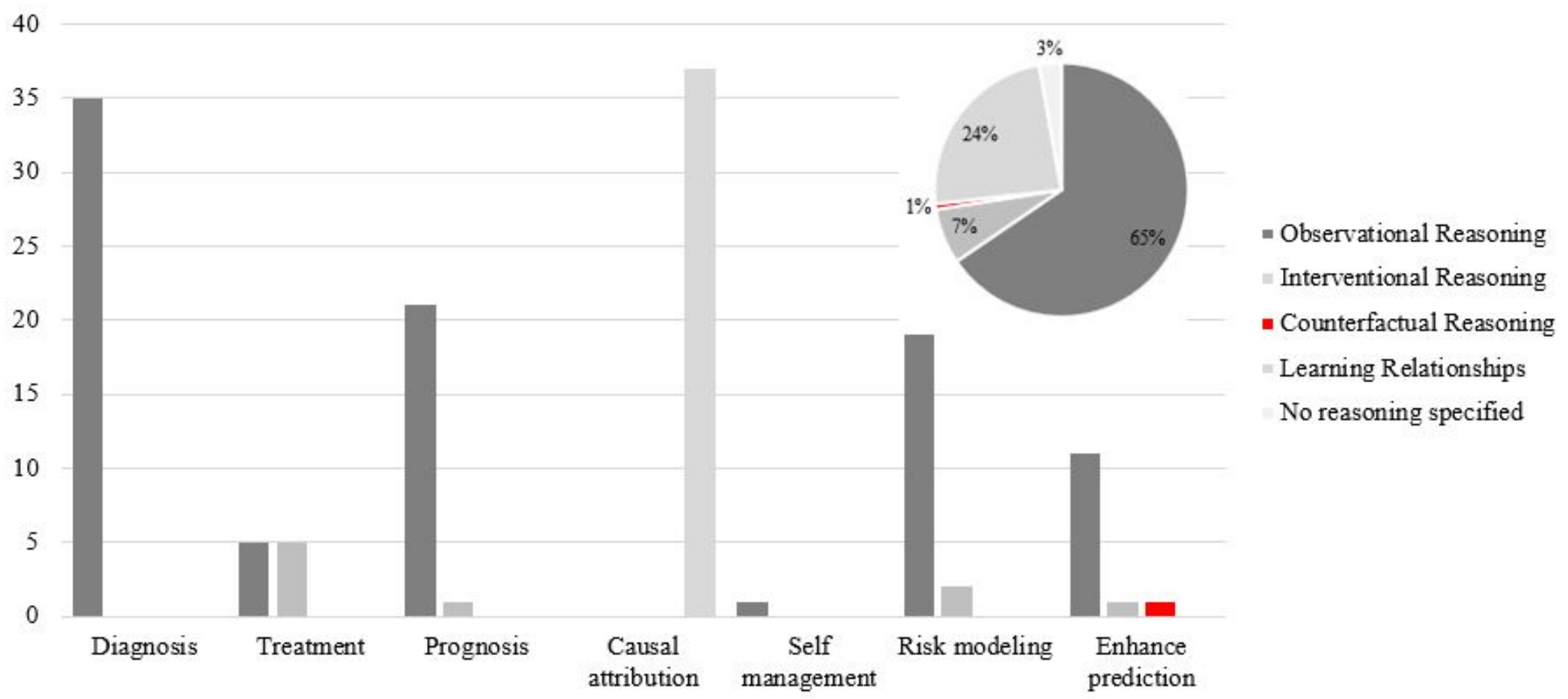

**Figure 5: Distribution of decision types and reasoning processes identified in BN literature.**

### 5.2.2 Modelling approach

Figure 6 presents the concept map of elements for the second objective of the review, whose main focus is the investigation of the BN development process with the goal of establishing current state-of-the-art from the literature. This section presents the results of the survey on this objective.

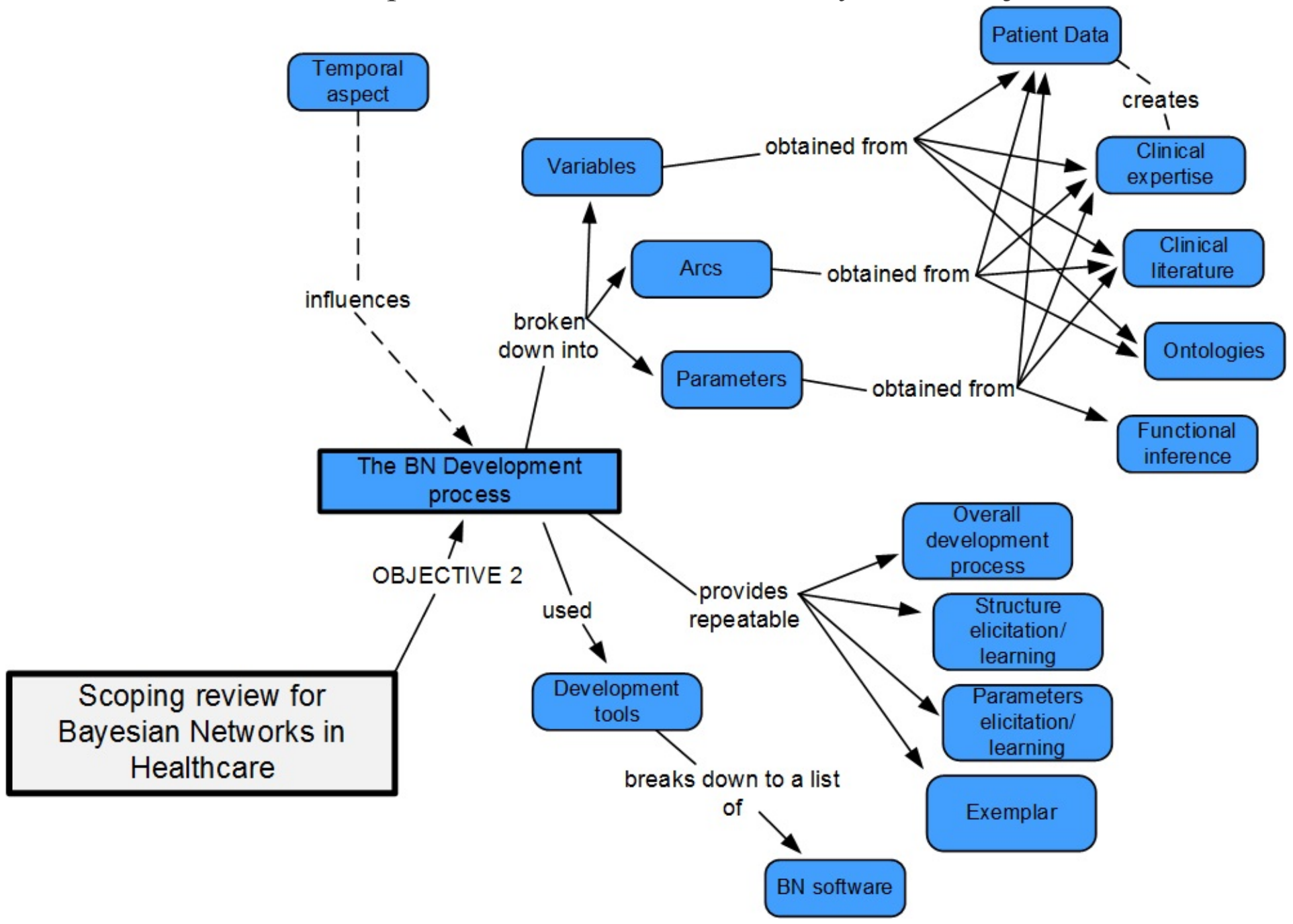

**Figure 6: Concept map for the second objective of the scoping review.**

The results on reviewing the *time dynamics* of proposed medical BNs are presented in Figure 7. Most papers described static BNs without a time element. The type of decision support possible influences, and is influenced, by the temporal aspect of the medical problem. Thus, there is an interconnection between Objective 1 and the time dynamics of the model. Static BNs, such as those presented by Vila-Frances et al. (2013), Farmer (2014), Khademi and Nedialkov (2015), Agarwal et al. (2016), Kuang et al. (2017), Ong et al. (2017) and Moreira and Namen (2018) were mainly proposed for diagnosis of a medical condition. A large number of static BNs were also applied to identify causal attributions, for example Liu et al. (2015), Zhang et al. (2015), Cardin et al. (2016), Soto-Ferrari, Prieto, and Munene (2017) and Gross et al. (2018). In other words, static BNs were used in situations where time was not a significant factor. In around 16% of the reviewed papers we identified BNs that implicitly captured time either through specific temporal nodes, like in the works of Akhtar and Utne (2014), Cai et al. (2015), Lee et al. (2016), Takenaka and Aono (2017) and Xu et al. (2018), or by modeling prior and/or post treatment variables such as those seen in the works of Constantiniou et al. (2015), (2016), Luo et al. (2017) and Sesen et al. (2013), (2014). In only 14% of the papers, for example the works of Sandri et al. (2014), Orphanou, Stassopoulou, and Keravnou (2016), Philipp et al. (2017) and Yousefi et al. (2018), temporal BNs were identified. While temporal BNs were mainly used in the performance of outcome prediction, a process described medically as prognosis (Exarchos et al. 2013; 2015; Velikova et al. 2014; Orphanou, Stassopoulou, and Keravnou 2014; Orphanou et al. 2016; Yousefi et al. 2018), we were also able to observe that prognosis models were quite uniformly distributed across the entire temporal scope. Finally, for almost 10% of the literature collection, including the works of Haddad, Himes, and Campbell (2014), Chang et al. (2015) and Vemulapalli et al. (2016) the presence of a time dynamic in the problem or proposed model was not clearly described and therefore classification of this element was not possible.

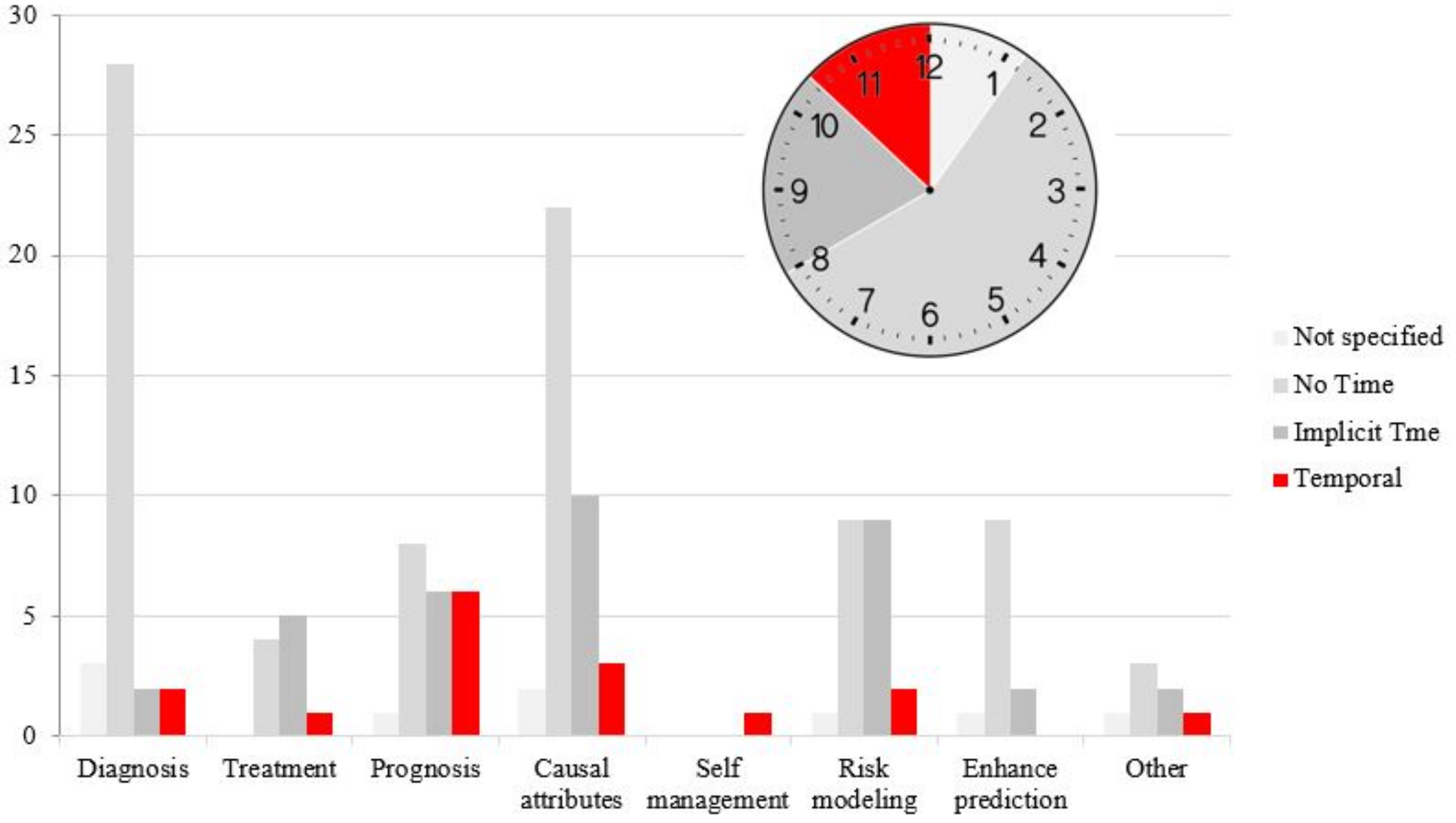

**Figure 7: Distribution of decision types and time dynamics identified in BN literature.**

As shown in Figure 8a, in most of the papers, including those of Alobaidi and Mahmood (2013), Loghmanpour, Druzdzel, and Antaki (2014), Khademi and Nedialkov (2015), Ojeme and Mbogho (2016), Liu et al. (2017) and Yetton et al. (2018), the *BN variables* originated purely from *data*. However, in almost 20% of the papers, the BN variables were obtained from *knowledge* elicited from medical experts or extracted from published scientific evidence. Further, in one-third of the knowledge-driven BNs such as those developed by Yet et al. (2014), Sandri et al. (2014), Kalet et al. (2015), Constantinou et al. (2016) and Philipp et al. (2017), the BN variables were elicited from experts through a series of interviews. In two papers published by Jochems et al. (2016) and Takenaka and Aono (2017) the variable selection was based on a previous study, while in another published by Chang et al. (2015), an ontology was used to resolve the BN structure. In few BNs, the variables were elicited either from both experts and literature (Yet et al. 2013; 2014; Bandyopadhyay et al. 2015; Farmer 2014; Velikova et al. 2014; Caillet et al. 2015; Yang et al. 2015; Tylman et al. 2016; Magrini, Luciani, and Stefanini 2018) or from a combination of clinical expertise and templates, such as ontologies and fragments (Constantinou et al. 2015; Refai et al. 2016; Marvasti, Yoruk, and Acar 2018). Finally, in almost 12% of the papers the BN variables originated from both data and knowledge, such as the works of Sesen et al. (2013), (2014), Frohlich et al. (2015), Neapolitan et al. (2016), Cypko et al. (2017), and Sa-ngamvang et al. (2018).

Referring now to *the arcs* that connect the variables in the BN, the majority were learned only from data using different constraint or score based learning algorithms (Figure 8b), such as those found in the works of Hernandez-Leal et al. (2013), Karaca et al. (2013), Liu et al. (2015), Vemulapalli et al. (2016), Jiao et al. (2017) and Aushev et al. (2018). In 25% of the published BNs, the arcs were elicited from knowledge, where clinical expertise was the main source, as in the work of Sandri et al. (2014), Zarikas et al. (2015), Constantinou et al. (2016), Phillip et al. (2017). In a paper published by Zarringhalam et al. (2014) the arcs between variables were elicited from literature, while in a smaller number of papers *rules*, described as arising from predefined *templates* or *ontologies*, were used (Vila-Francés et al. 2013; Akhtar and Utne 2014; Chang et al. 2015; Agarwal et al. 2016; Gámez-Pozo et al. 2017). In addition, a combination of *knowledge and rules*, specified as templates and *idioms* (Neil, Fenton, and Nielsen 2000) was also used to elicit the BN arcs, such as in Yet et al. (2013), (2014), Velikova et al. (2014), Orphanou et al. (2016), Magrini et al. (2018). Furthermore, *mixed approaches*, where structure learning algorithms alongside causal constraints provided by experts, literature or causal rules, were used, for example in the works of Fuster-Parra et al. (2016), Chao et al. (2017), Sesen et al. (2014), Shen et al. (2018) and Xu et al. (2018).

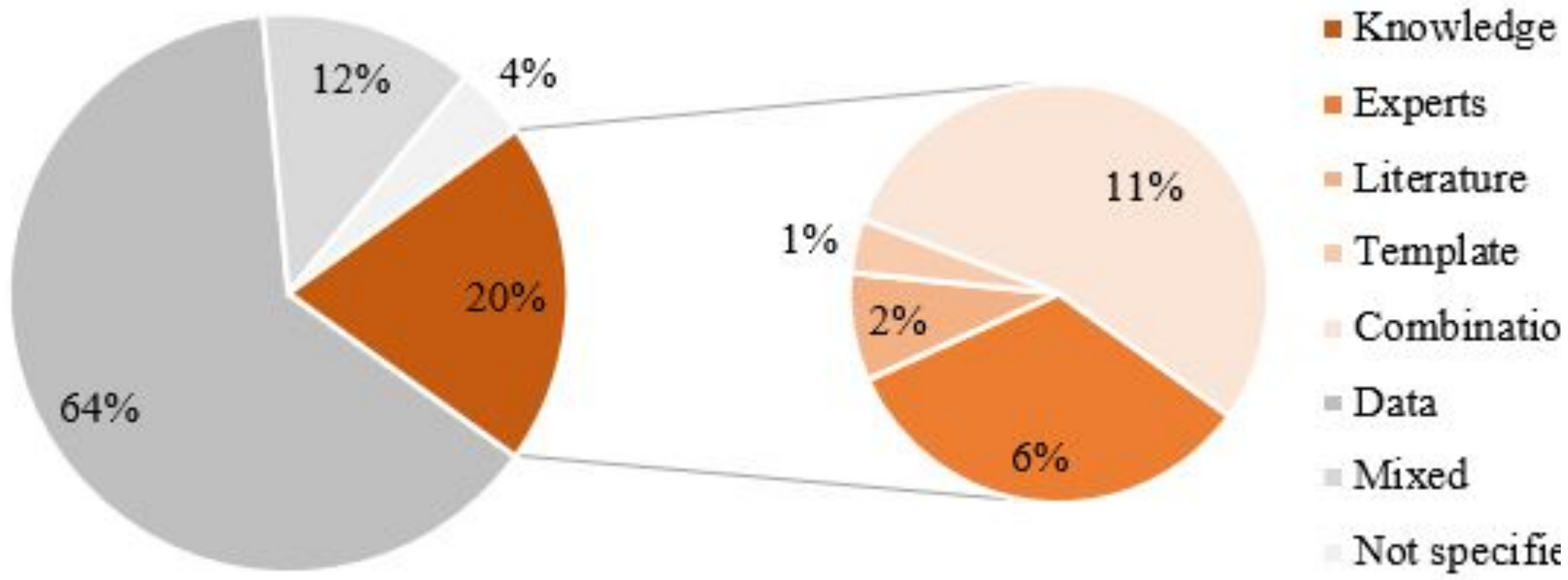

a)

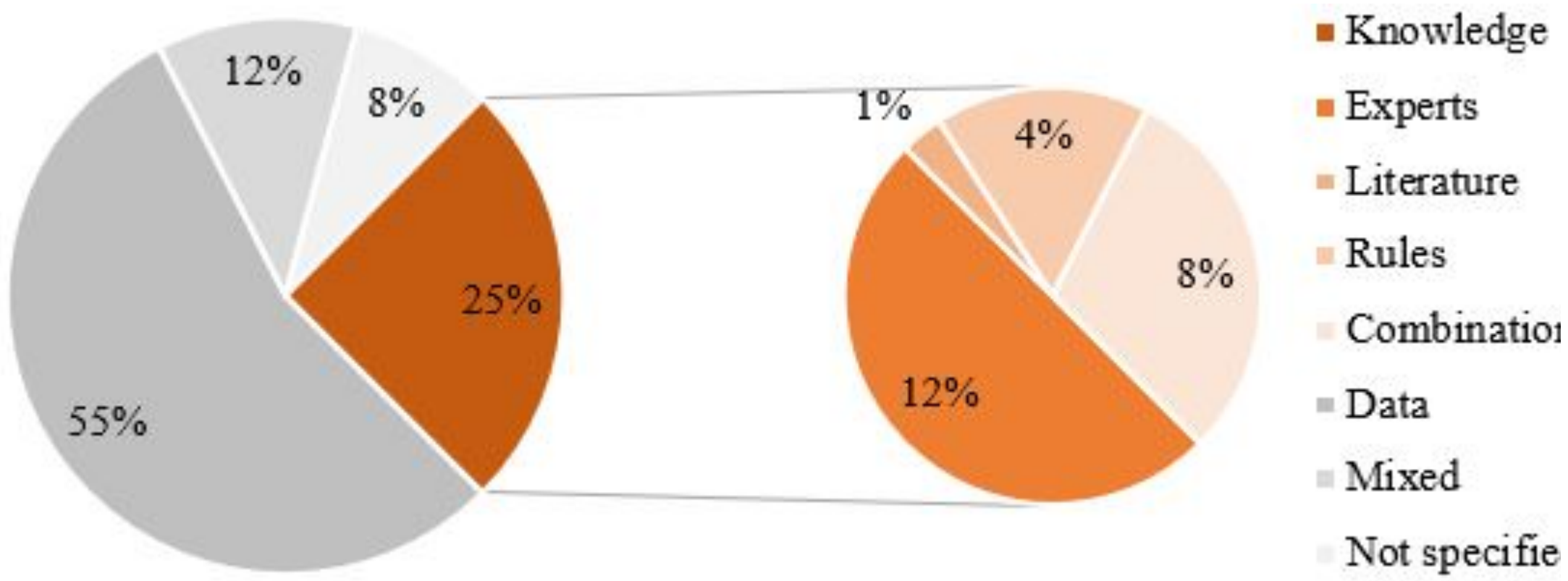

b)

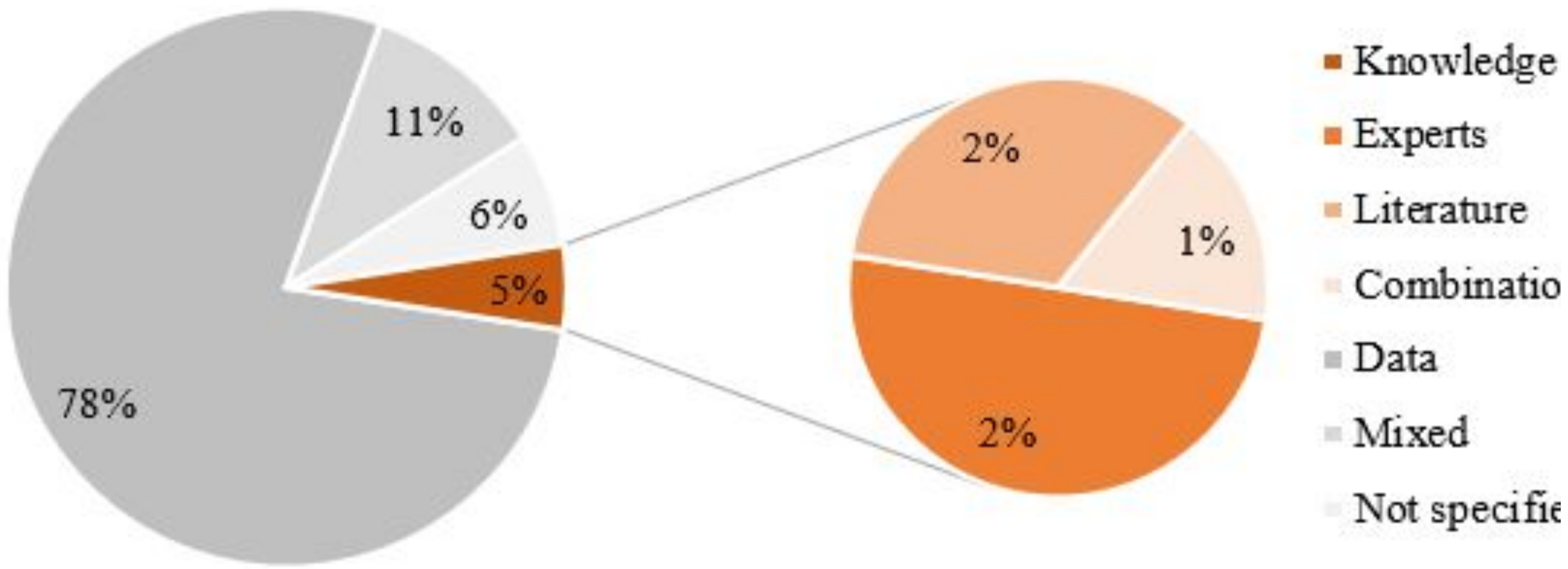

c)

**Figure 8: Origin of BN inputs; a) variables, b) arcs, c) parameters.**

As shown in Figure 8c, in the vast majority of papers, including those Gianaroli et al. (2013), Akhtar and Utne (2014), Khademi and Nedialkov (2015), Lee et al. (2016), Nandra et al. (2017) and Aushev et al. (2018), the *BN parameters* were learned purely from *data*. In only 5% of the papers the parameters were learned from *knowledge* with clinical expertise being the main source of it (Agarwal et al. 2016; Lou et al. 2017; Philipp et al. 2017). In two papers the BN parameters were elicited only from literature (Moreira et al. 2016; Tylman et al. 2016), while in another one the parameters were elicited from a combination of experts, literature and the use of mathematical functions (Velikova et al. 2014). In 11% of the published BNs their parameters were based on both data and knowledge, like in the works of Yet et al. (2014), Constantinou et al. (2015), (2016), Liu et al. (2017), and Magrini et al. (2018).

Each paper was evaluated regarding how detailed the overall *BN development process* was. In 48% of the studied papers, for example Berchialla et al. (2014), Yang et al. (2015), Zeng et al. (2016), Alobaidi and Mahmood (2013) and Xu et al. (2018), the steps followed for developing and validating the BN were either completely omitted or could only be implied. In less than a quarter of the literature, the overall development process was either described in detail in the text, for instance in the works of Kalet et al. (2015) and Magrini et al. (2018), or visually illustrated as seen in the works of Yet et al. (2014), Akhtar and Utne (2014), Constantinou et al. (2015), Chao et al. (2017) and Park, Chang, and Nam (2018). *Generally, in this study we found that papers written by decision scientists and clinicians collaboratively tended to be more repeatable than those papers written by either group alone*.

Apart from analyzing the overall development process, we further investigated whether the *BN structure and parameter learning*, *or elicitation process* was detailed enough to be considered as repeatable. As shown in the inner pie of the doughnut graph in Figure 9a, from the papers where the origin of the BN structure was specified, 65% were learned from data (either completely or partially) and 35% from knowledge (either completely or partially). The outer layer of Figure 9a identifies that in situations where the BN structure was learned from data, papers were almost uniformly distributed between those with negligible, insufficient, and sufficient detail regarding the structured learning process. In 24% of the papers, such as those written by Lappenschaar et al. (2013), Berchialla et al. (2014), Yang et al. (2015), Turhan, Bostanci, and Bozkurt (2016), Bukhanov et al. (2017) and Seixas et al. (2018), the authors gave only simple statements such as *the BN structure was learned from data or the structure was learned using supervised (or unsupervised) learning algorithms*, which are incomplete descriptions that do not allow the reader to fully understand the process used. In 23% of the data-driven BNs greater detail, such as the name of the learning algorithm used, was provided. Examples include: Exarchos et al. (2013), Cai et al. (2015), Zador, Sperrin, and King (2016), Jiao et al. (2017) and Liu et al. (2018). Finally, in 18% of the data-driven BNs a detailed description of the structured learning approach was provided, such as those found in Ji and Wang (2014), Fuster-Parra et al. (2016), Mcheick et al. (2017) and Hu and Kerschberg (2018). Conversely, in the vast majority of papers where the BN structure was elicited from knowledge no description of the elicitation process was given, as in the works of Bandyopadyay et al. (2015), Wang et al. (2015), Orphanou et al. (2016), Cypko et al. (2017),

Xu et al. (2018). Statements such as *the model was constructed by two experts or structure was based on expert domain knowledge or the structure was determined by combining known relationships from the medical literature with input from our clinical collaborations* provided the totality of methodological description. In situations where the BN structure was derived from ontologies, the process was usually described with sufficient detail (Akhtar and Utne 2014), (Chang et al. 2015), (Agarwal et al. 2016). In two papers published by Kalet et al. (2015) and Seixas et al. (2014) a detailed description of the elicitation process was provided. Regardless of the origin of the BN structure, a complete BN was provided in 53% of the papers, while a description of the variables included in the BN and their states were provided in only 28% of the studied papers, such as Sesen et al. (2013), Qian et al. (2014), Constantinou et al. (2015), Orphanou et al. (2016), Nandra et al. (2017), Moreira and Namen (2018).

As shown in the inner pie of the doughnut graph in Figure 9b, from the papers where the origin of the BN parameters was specified, 85% were learned from data (either completely or partially) and 15% from knowledge (either completely or partially). In 58% of the papers, where BN parameters were learned from data, no description of the learning process was provided, as in Hernadez-Leal et al. (2013), Gianaroli et al. (2013), Liu et al. (2015), Frohlich et al. (2015), Refai et al. (2016), Ong et al. (2017), Park et al. (2018). In such cases, the authors simply mention that the BN parameters were learned from data without any further information. Only three papers provided a detailed description of the parameter learning process (Ducher et al. 2013; Van der Heijden, Velikova, and Lucas 2014; Constantinou,et al. 2015). Regarding the knowledge driven BN parameters, in 10% of the papers, the parameter elicitation process was not described, for example Tylman et al. (2016) and Lou et al. (2017). Finally, in only three papers the parameter elicitation process was described in depth (Yet et al. 2013; 2014; Constantinou et al. 2015). Regardless of the origin of the BN parameters, marginal probabilities or complete NPTs were rarely provided. More specifically, in 27% of the papers some parameters were presented, while a full list of parameters was available in only 7 papers (Karaca et al. 2013; Qian et al. 2014; Jochems et al. 2016; Lee et al. 2016; Neapolitan et al. 2016; Dranca et al. 2018; Merli et al. 2016).

a)

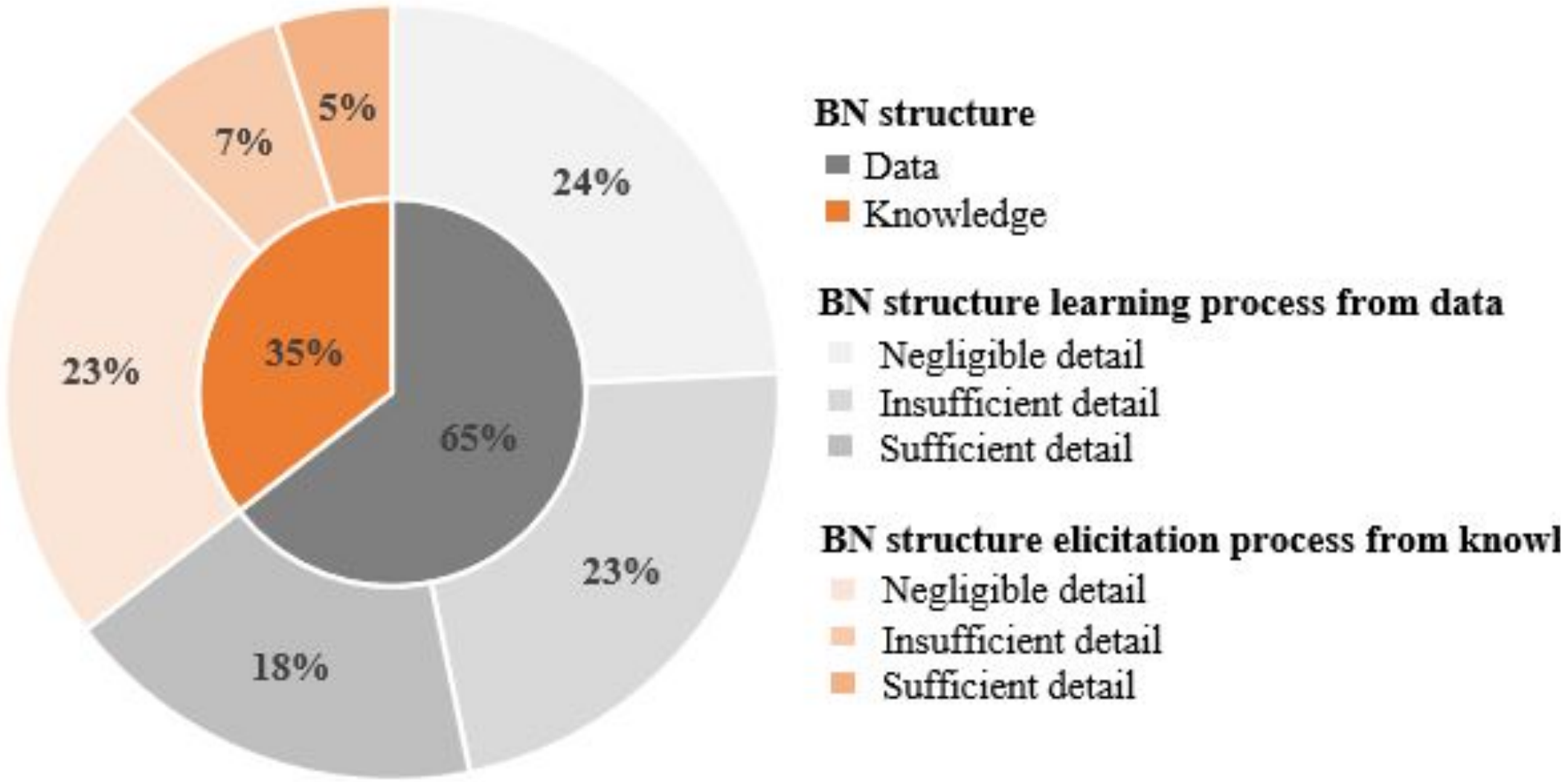

b)

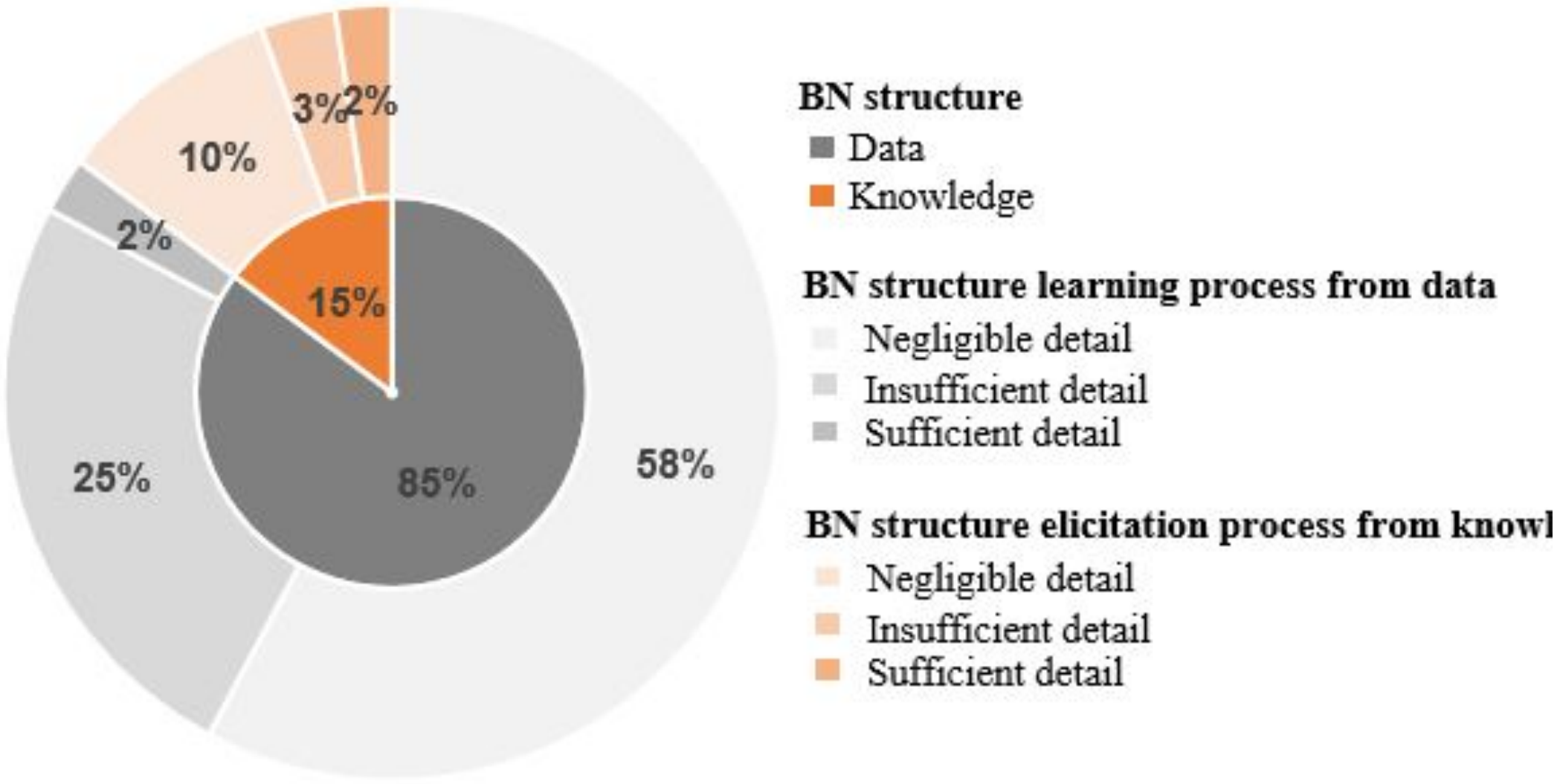

**Figure 9: Detailed BN development process; a) BN structure learning and elicitation process, b) BN parameter learning and elicitation process.**

The last modelling approach element studied in this part of the review was to identify the *BN software* used by authors. We see from Figure 10 that in 35% of the papers either the software used was not specified or the authors implemented their own ad-hoc software (Hernandez-Leal et al. 2013; Exarchos et al. 2013; Haddad et al. 2014; Li et al. 2015; Moreira et al. 2016; Ong et al. 2017; Shen et al. 2018). The two software tools most often used were GeNIe™ (Jochems et al. 2016; Cypko et al. 2017; Magrini et al. 2018) and R™ (Zador et al. 2016; McNally et al. 2017; Noyes et al. 2018).

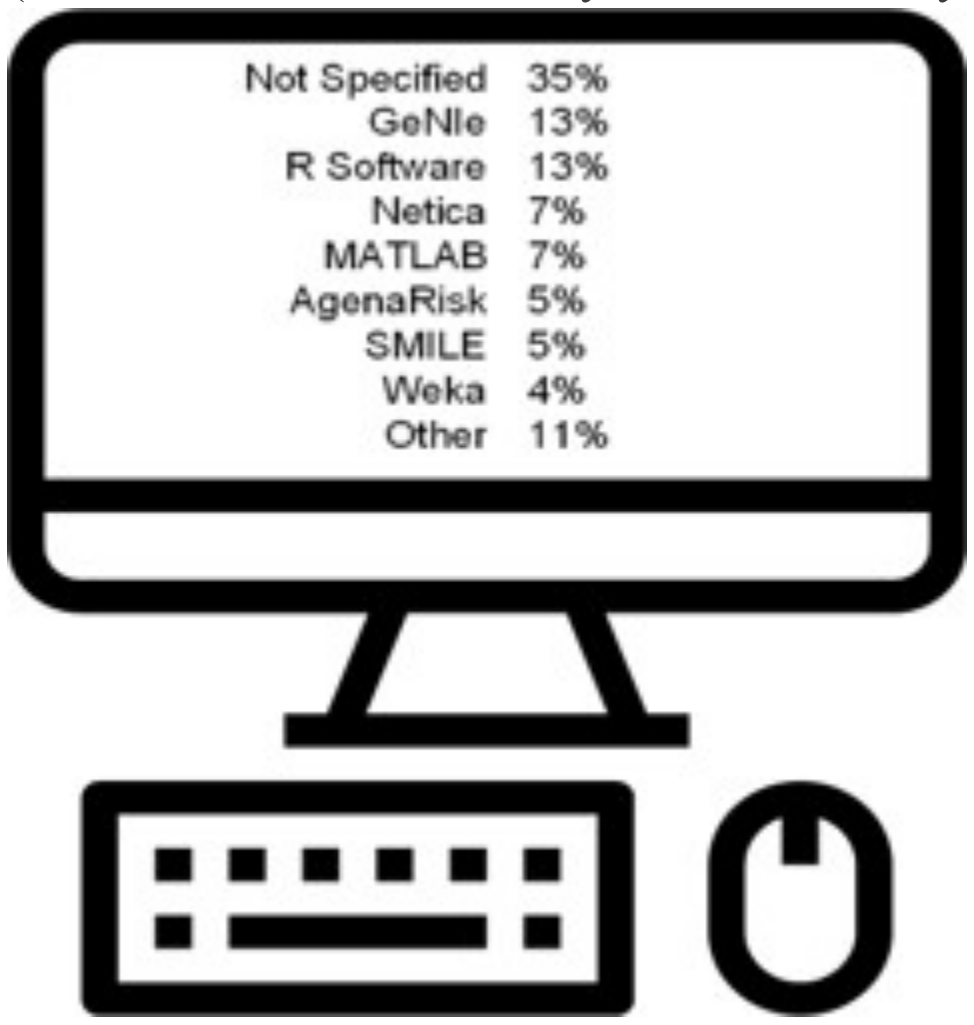

**Figure 10: Software used in BN Development.**

### 5.2.3 Model usefulness

Figure 11 presents the concept map of elements for the third objective of the scoping review, whose focus is to identify the BN usefulness in clinical practice. This section presents the results of the survey on this objective.

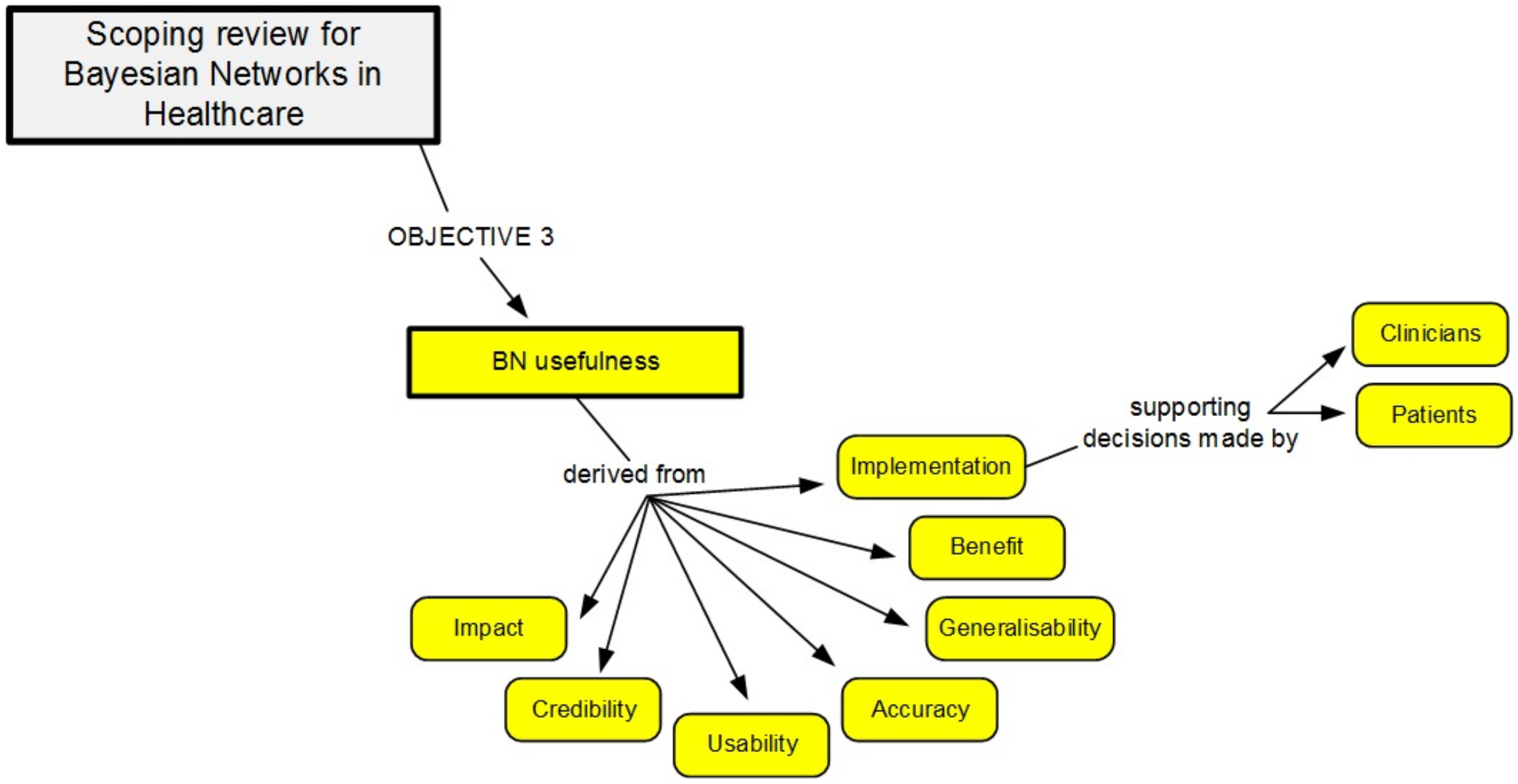

**Figure 11: Concept map for the third objective of the scoping review.**

Figure 12 presents the frequency of elements related to BN usefulness in healthcare. Most papers clearly described some *clinical benefit* for their medical BN, as in Berchialla et al. (2014), Yet et al. (2013), Velikova et al. (2014), Li et al. (2015), Agarwal et al. (2016), Nandra et al. (2017), Yousefi et al. (2018). Many papers also provided BNs where an accepted approach was used to evaluate and report the *accuracy* of models and their predictions, like in the works of Velikova et al. (2013), Sandri et al. (2014), Yin et al. (2015), Zador et al. (2016), Kuang et al. (2017), Shen et al. (2018). However, *generalizability* could be identified in only 5% of papers, for example: Li et al. (2015), Orphanou et al. (2016), Soto-Ferrari et al. (2017) and Luo et al. (2018). From the published BNs, only 34% appeared to be *credible* per the definition in Table 1. Two thirds of these were knowledge-driven, a strong indicator that the BN structure was sensible, e.g.: Yet et al. (2014), Zarikas et al. (2015), Constantinou et al. (2016), Gamez-Pozo et al. (2017) and Magrini et al. (2018). From the credible BNs only 8 were learned completely from data, including those of: Lappenschaar et al. (2013), van der Heijden et al. (2014), Jin et al. (2016), Mcheick et al. (2017) and Ong et al. (2017). In the majority of data-driven BN structures, the model's credibility was difficult to be identified. Two of the rarest characteristics to identify were the model's *usability* and *impact*. To find usability we sought whether the authors proposed or demonstrated an interface, as well as whether they had also validated that interface using experts. In less than 1% of the papers an interface was proposed. Examples can be found in Vila-Frances et al. (2013), Liu et al. (2015), Chang et al. (2015), Agarwal et al. (2016), Mcheick et al. (2017), Park et al. (2018). From the papers, where an interface was presented, only one described evaluation of the proposed interface (Chang et al. 2015). Two types of *impact* were identified: (1) *potential impact*, indicating that a small study on whether the developed BN can impact clinical decision making was performed; and (2) *actual impact*, indicating that a trial had been conducted to evaluate whether the BN could be applied in a longer-term implementation. In only three papers the *potential impact* of the developed BN was investigated (Constantinou et al.2015; Chao et al. 2017; Mcheick et al. 2017). The *actual impact* of the BN was not identified in any paper in this survey. *Finally, there was nothing to suggest that any of the BN models in this review were in current clinical use.*

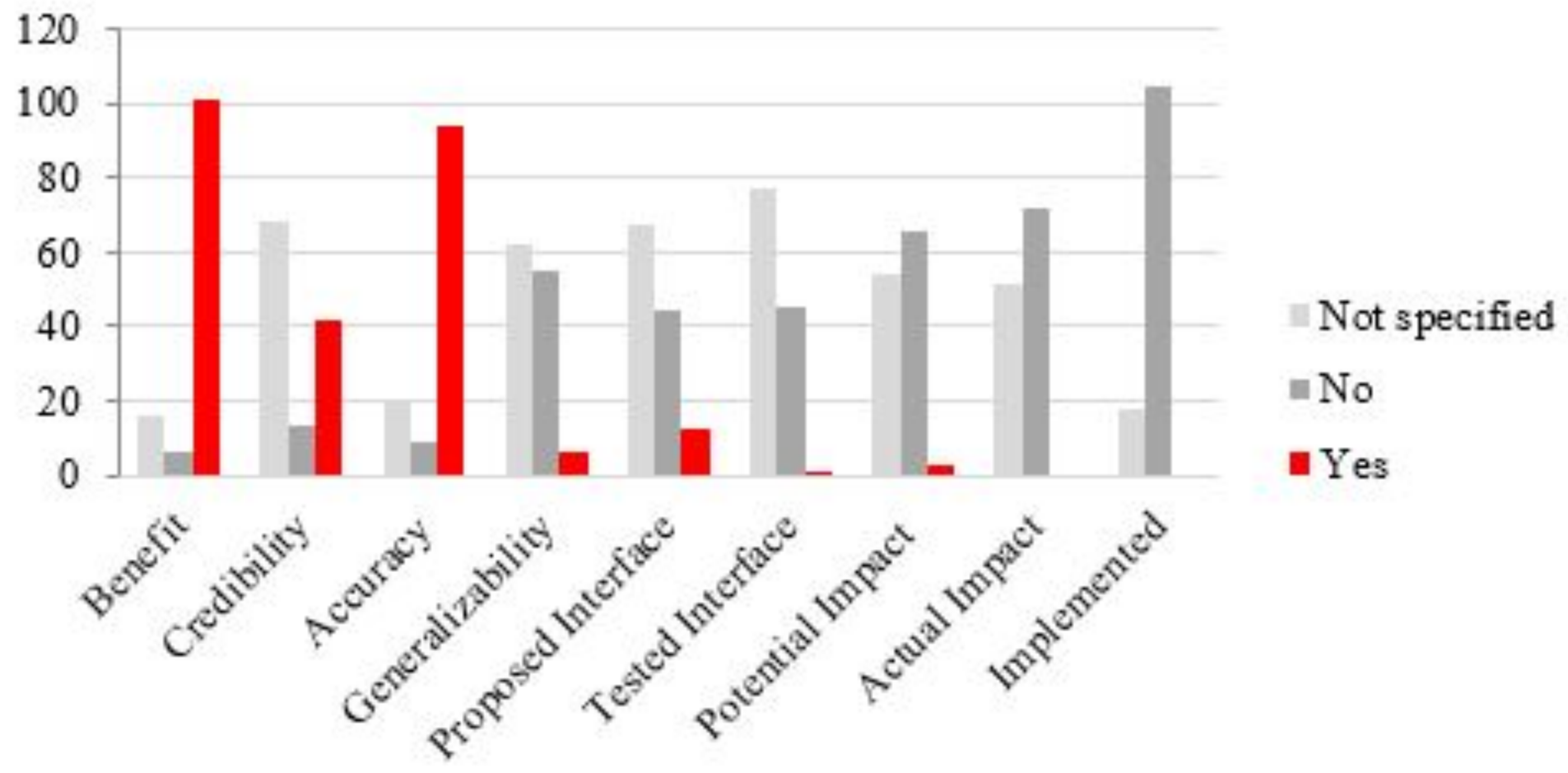

**Figure 12: Frequency of the properties related to the BN usefulness in healthcare.**

# 6 Major Findings

During the review a taxonomy was deductively resolved that classifies concepts related to each of the three objectives. Various approaches exist for resolving knowledge and defining the taxonomy, but at their core taxonomies are systematic classifications within a domain of interest that typically present as hierarchical, multidimensional arrangements of linked categories of concepts (Gilchrist 2003; Kakabadse, Kakabadse, and Kouzmin 2003; Bowman and Nichols 2002). It is not unusual to observe taxonomies represented in almost every conceivable shape, including: linear (Triplett 2018), circular (Atkinson 2015), flow diagrams (Van Der Maaten, Postma, and Van Den Herik 2009) and so on (Jordan 2007). While most taxonomies are observed to be unidirectional and therefore arranged similar to a concept hierarchy, from most general to most specific, our review resolved a bi-directional taxonomy for each of the three objectives presented in this work. Each taxonomy also provides secondary information in that the weight of the connecting line, as described in Figure 16, represents the frequency of papers (as an overall percentage of papers in the review) whose work is represented by the relationship between the two identified nodes.

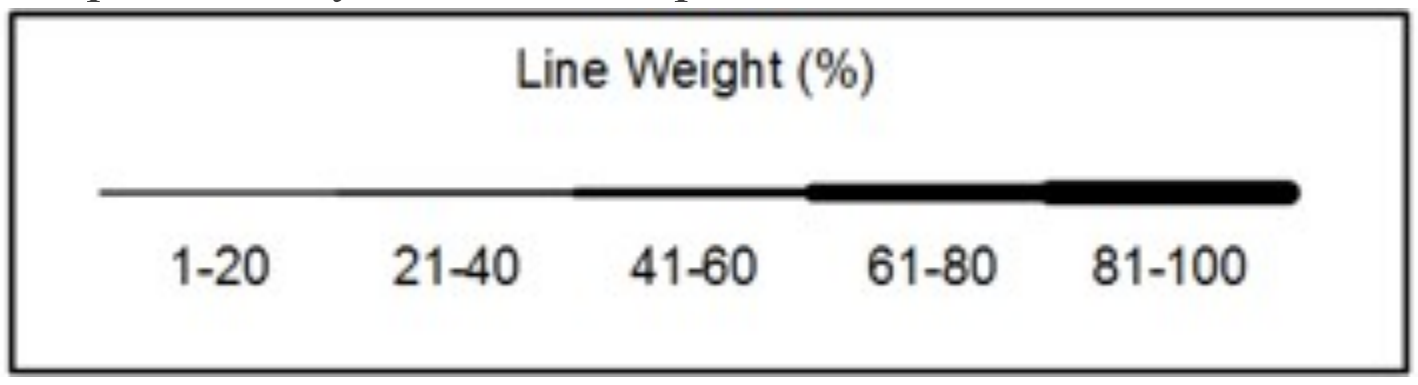

**Figure 16: Line weight and percentage of papers.**

## 6.1 Decision support

The first taxonomy presented in Figure 17 contextualizes the relationship between the *problem to be solved* and the type of *decision support* being employed. Three *reasoning types* were identified from the literature. As can be seen by the indicated line weight, the vast majority of developed BNs engaged in a process of *observational* reasoning, leaving *interventional* and *counterfactual* reasoning significantly underexplored. Further, four *clinical decision types* and four *decision contexts* were identified. *Diagnosis* was the most frequent clinical decision, for which only reasoning from evidence was required. The most frequent decisions related to *interventional reasoning* were *treatment recommendations*. From the decision contexts, learning *causal attributes* was more frequently observed. A summary of the major findings related to the first objective of this scoping review is presented in Textbox 1. A further detailed discussion on these findings can be found in Section 7.

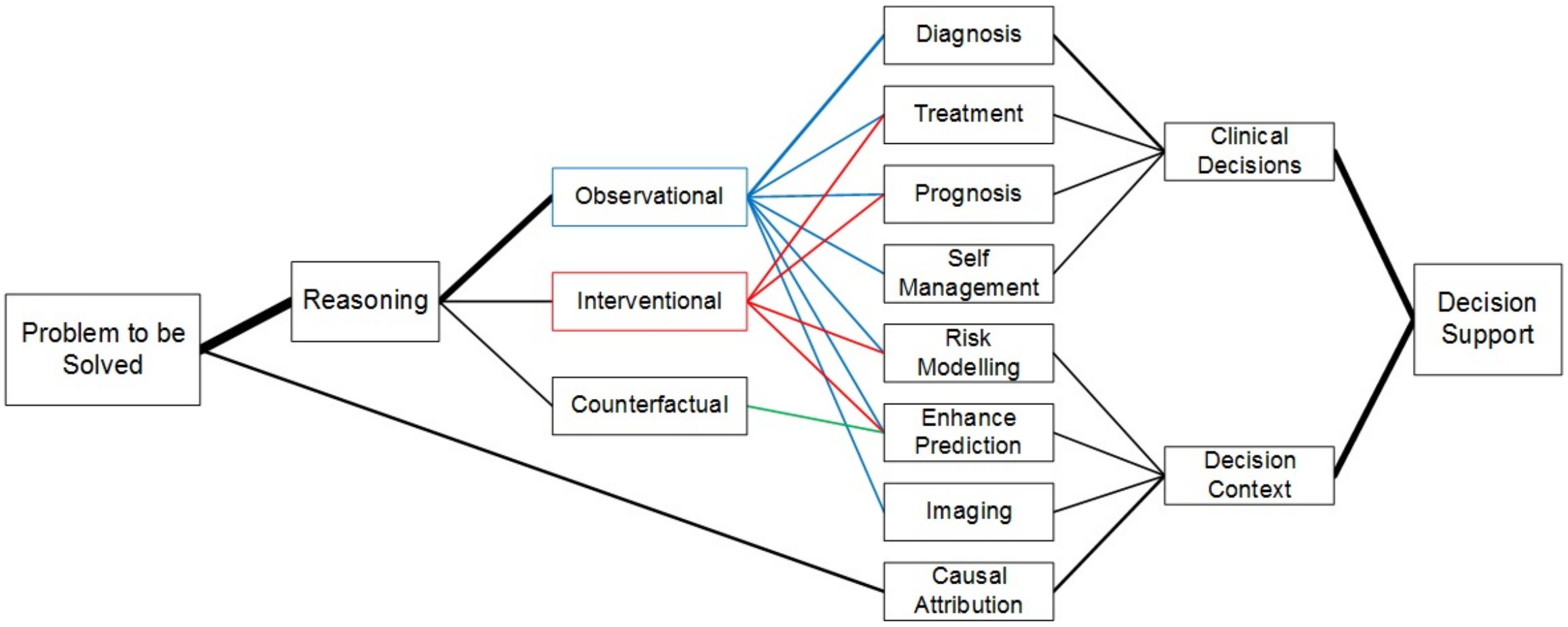

**Figure 17: Taxonomy of the first objective of the scoping review.**

**Textbox 1: Major findings related to the first objective - decision support.**

1. Majority of developed BNs support reasoning from evidence
2. Few BNs on interventional reasoning
3. Only one BN on counterfactual reasoning
4. The two most common aims of the published BNs were to help diagnosis and learn causal attributes from data

## 6.2 Modeling approach

The second taxonomy shown in Figure 18 focuses on the interplay between *model development* and the *developers* engaged to the task. Three types of *time dynamics* were identified with more than half of the published BNs identified as *static*: used in circumstances where no time element was essential to the model. Three elements, *variables*, *arcs* and *parameters*, are necessary for representing the developed BN. As can be seen in the taxonomy, *data-driven* approaches were more common than the *knowledge-driven* methods that rely on *experts, literature or ontologies*. The approach of learning, or eliciting, each element was characterized as being *bespoke* to the published case, or *generic* to the degree of being potentially repeatable. On the *developer* side our review identified two primary *developer* actors: *decision scientists*, such as computer scientists and statisticians; and *clinicians*. We were able to classify three potential development approaches: *method-, problem-* and *hybrid-* driven. It was apparent from a thorough reading of the literature that *hybrid-driven* papers provided a more generic development approach. A summary of the major findings related to the second objective of this scoping review, is presented in Textbox 2. A further detailed discussion on these findings can be found in Section 7.

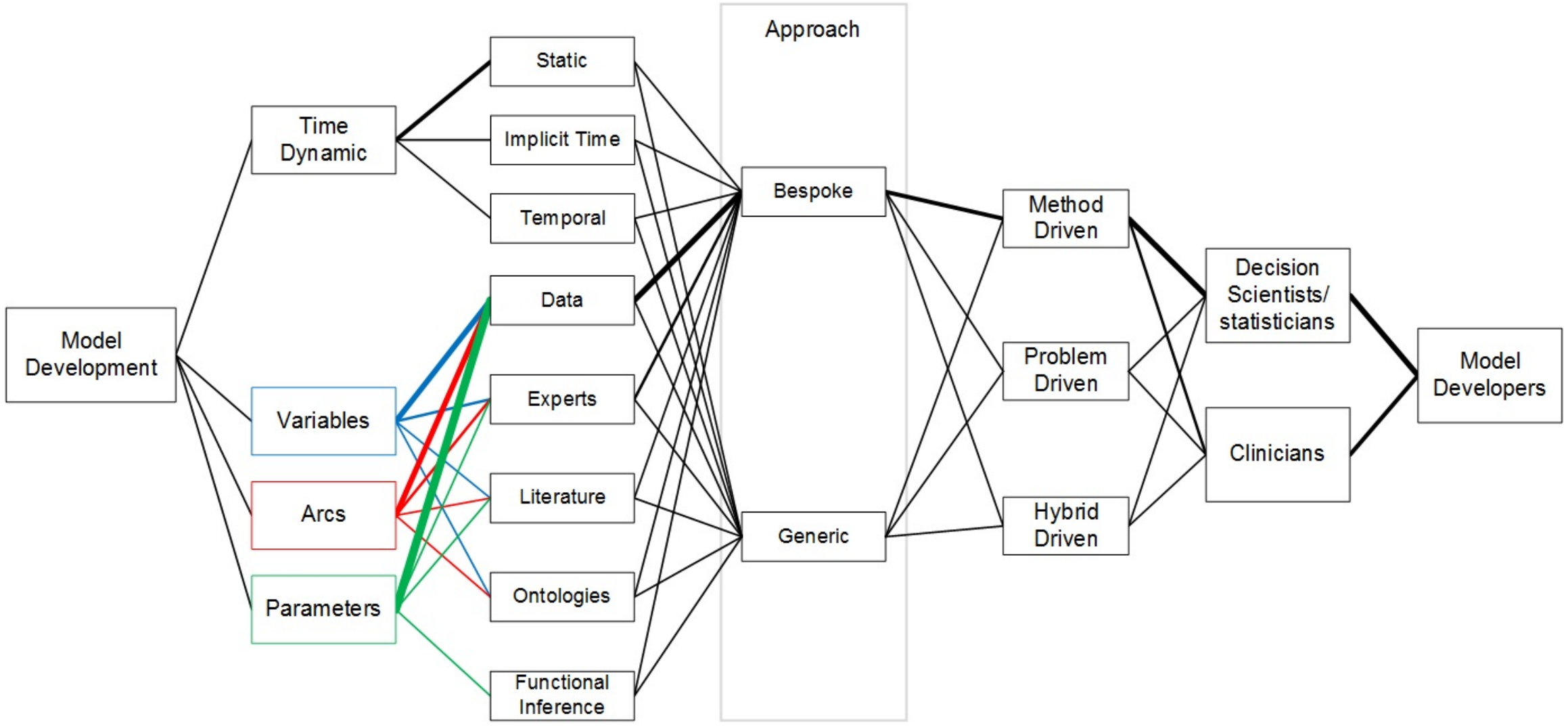

**Figure 18: Taxonomy of the second objective of the scoping review.**

**Textbox 2: Major findings related to the second objective - modeling approach.**

1. 57% of the published BNs were static models with no time element
2. A prevalence of *data-driven* over *knowledge-driven* models
3. The methodologies employed for learning *data-driven* BNs was described in greater detail than that for eliciting *knowledge-driven* BNs
4. The process of eliciting the BN structure and/or parameters from experts was rarely described in the literature
5. Experts were a major source of knowledge for eliciting the BN structure and/or parameters
6. The literature lacked a well described overall development process
7. *Hybrid-driven* papers were found to be more generic in approach and application
8. The complete BN structure was not provided in 47% of the papers reviewed and in those where the BN structure was available, a detailed description of the variables was rarely given
9. NPTs were not provided in 73% of papers in this review

## 6.3 Model usefulness

Figure 19 presents the third and final taxonomy, which classifies the BN's potential or claimed usefulness and connects this with elements that can measure potential for successful engagement of BNs in clinical practice. Six main elements were identified for the potentially useful BN that are grouped into three main categories. These evaluate the BN's: *applicability, validity* and *adoptability*. As can be observed from the thickness of the connecting lines, *benefit, accuracy*, and *credibility* were the most observed elements in this review. It was interesting to note that the BN model's *adoptability* was identified as the most neglected category, as this was also reflective of the lack of works claiming to present a model that had been adopted in some way into clinical practice. A summary of the major findings related to the third objective of this scoping review, is presented in Textbox 3. A further detailed discussion on these findings can be found in Section 7.

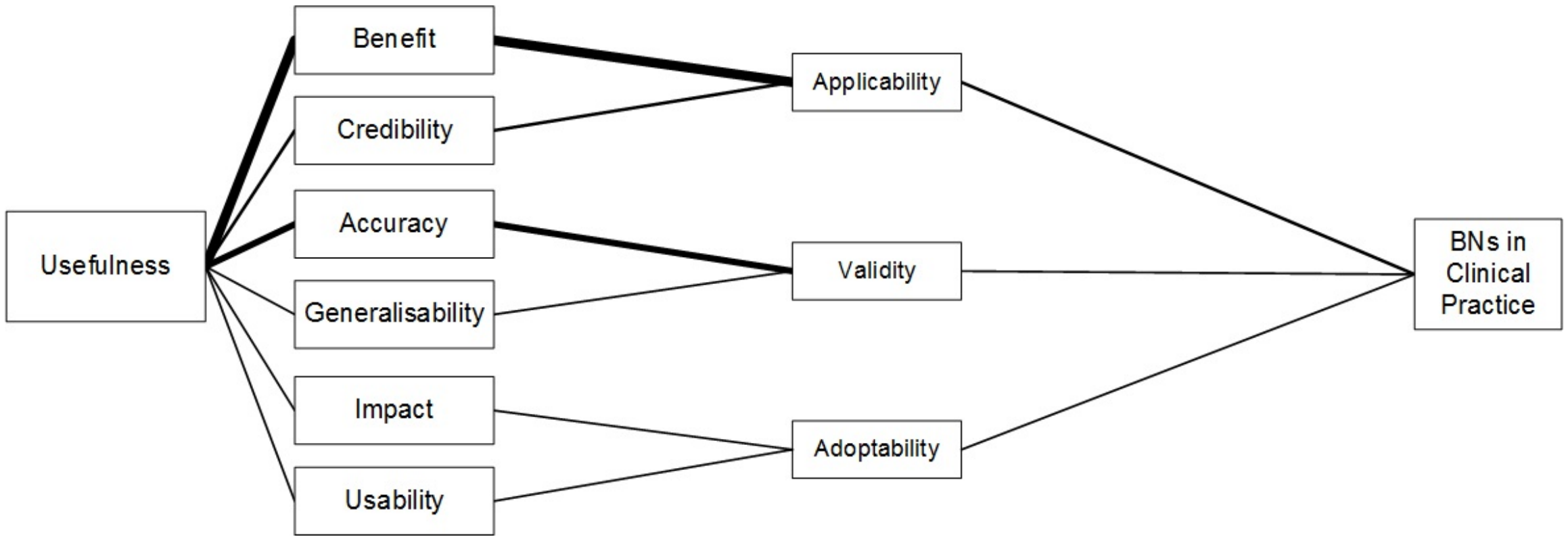

**Figure 19: Taxonomy of the third objective of the scoping review.**

**Textbox 3: Major findings related to the third objective - model usefulness.**

1. The elements most frequently identified in the literature were benefit, credibility and accuracy
2. *Knowledge-driven* BNs appeared more credible on review than *data-driven* BNs
3. A model's *generalizability* and *usability* were rarely specified
4. The clinical *impact* of developed BNs was the most underexplored element in the literature
5. A gap exists between *developing an accurate model* and *having a useful model that can be adopted in healthcare*

# 7 Major Findings

This scoping review intended to examine and evaluate the: (1) targeted *decision support*; (2) *modelling approach*; and (3) *usefulness* in practice for BNs in healthcare. We found that the literature on BNs in healthcare could be broadly grouped into: (1) *method-driven*, (2) *problem-driven*, and (3) *hybrid-driven* papers. On one hand, *method driven papers* were seen to be primarily written by decision scientists and their focus was on a specific aspect of the modelling methodology. Hence, the BN *development* and *validation* are usually described in greater detail than the *clinical benefit* and *potential impact* in practice, which appear to have been neglected by most researchers. On the other hand, *problem driven papers* were mainly written by clinicians and their focus is on the medical application. In this type of papers, the *clinical benefit* and *medical background* are usually described in detail, while the BN *development* is usually only vaguely explained. *Hybrid papers* result from a collaboration between decision scientists and clinicians and the content of their works tended to be more equally divided between the clinical and modelling aspects of the BN. Hybrid papers were found to be the most complete of the three types of papers. The remainder of this section provides a more thorough analysis of the review findings associated with each objective, while also discussing the strengths and weaknesses of the review process undertaken.

## 7.1 Discussion of findings

**Decision Support:** In many of the published BNs no real reasoning was performed as the aim of most works was to identify, or learn, relationships or causal attributes from the data. In BNs where reasoning was undertaken, the vast majority sought to help clinical decision-making, such as for diagnosis or prognosis. In these instances, reasoning from evidence was most often used. *Interventional* and more especially *counterfactual reasoning* were significantly underexplored, indicating that BNs are not being used to their full potential. Despite the benefit of counterfactual reasoning in answering important questions regularly asked by clinicians, such as *Would the patient have survived, if we had operated him sooner?*, counterfactual analysis is still considered by many as a contentious research approach. There is a debate as to whether counterfactual reasoning has scientific validity, or whether it is merely a metaphysical process with no observable or testable consequences (Dawid 2000). This might explain the extremely small number of BNs in which a counterfactual reasoning approach was undertaken. While most of the published BNs are static models with no or implicit time element, we know that time plays a crucial role in clinical decision-making and that BNs can model these dynamic processes yet temporal BNs remain less popular. This might be explained by the difficulty of modelling temporal BNs, as they are structurally more complex and require more computational power to perform inference. It may also be due to the perceived unavailability or incompleteness of temporal datasets from which to draw knowledge necessary to making those inferences.

**Modeling Approach:** In 48% of the papers, description of the *overall BN development process* was either absent or could only be inferred from the description of the case study. In only a small number of papers was the BN development description detailed enough to reach a level of being potentially **repeatable**. The number of BNs where the structure was learned only from data was twice that of BN structures elicited from knowledge. This may be because structured learning is less time-consuming and is usually based on well-known learning algorithms, which makes the process more repeatable. The vast number of data-driven BNs in medical applications came as a surprise if we consider the fact that they are based solely on medical data, which are not always good enough or complete enough to be used as the only source for learning a causal structure (Hernandez-Leal et al. 2013; Magrini et al. 2018). Many times, the quality of medical data for research purposes was described as poor, and poor data therefore means meaningless data-driven BN structures (Constantinou et al. 2016). Moreover, even if the available data includes information on all necessary variables, there are typically far too many variables and far too few samples in the dataset to achieve any sensible structural learning outcome (Constantinou et al. 2016). When the BN structure is learned from knowledge, the structure is more sensible as it can more correctly represent the true causal mechanisms. However, the elicitation process is time-consuming and may be prone to errors, especially when it is based only on the input of a single domain expert (Bielza and Larranaga 2014; Constantinou et al. 2015). Additionally, the way a *BN structure* was elicited from knowledge was rarely described in the literature, making the process ad hoc and less able to be standardized. In other words, elicitation process descriptions were found to be ambiguous as insufficient detail was provided for the reader to understand where the BN structure came from. This makes it near impossible to duplicate the authors' process and duplicate their findings. Regardless of the origin of the BN structure, in almost half of the papers the complete BN structure was not provided. This also makes the process of verifying or using the structure impossible. In 85% of the papers the *BN parameters* were learned from data (either completely or partially). The process, either data or knowledge driven, was usually not described in detail. Parameter learning from data is more straightforward, while eliciting parameters from knowledge is less intuitive and results in subjective parameters, therefore a detailed description of the parameter elicitation should be a requirement of any BN in healthcare publication. Regardless of the origin of the parameters, the NPTs, which could be used as an important source of knowledge for other researchers, were rarely provided in the literature. Finally, in spite of the advantages of mixed approaches for developing BN structures and parameters using both data and knowledge, only a small number of BNs developed by mixed approaches were identified in the literature.

**Model Usefulness:** While the majority of published BNs addressed a clinical problem of significant importance and were shown to have an acceptable predictive performance, there appeared a significant gap between *having an accurate model* and *demonstrating a clinically useful model capable of delivering an actual impact on healthcare decision making*. A description of the model's credibility was frequently observed in the literature. Particularly, papers with knowledge-driven BNs provided more convincing evidence of model credibility than those with data-driven BNs. Important elements for having a useful model, such as *generalizability*, *usability* and *impact* were either vague or overlooked in the literature. Generalizability is important for achieving BN adoption in healthcare as a model could not be used widely in practice if it has not first been proven to work on alternative populations and therefore applied to the broader health consumer population. Another important aspect considered was whether user-friendly interface and an impact analysis to study the benefit of the model and its integration into the existing workflow had been performed. In this review, no single BN adhered to all of the prescribed elements necessary for having a useful model. In addition, there was no evidence in this review to suggest that any of the BN models were in current clinical use. This does not categorically mean there are no BNs in clinical practice, only that in this review we did not find a publication describing its BN as currently engaged in clinical practice.

## 7.2 Strengths and limitations

The first strength of this review is that this is the largest review of BNs in healthcare targeting three objectives; (1) *decision support*, (2) *development*, and (3) *usefulness*, which, when combined, provide a comprehensive assessment of the domain. Several attributes related to each objective were identified based on a *preliminary short review* of the literature (Kyrimi et al. 2020) followed by *a comprehensive review* process conducted by a panel of experts. The second strength is that in order to ensure a consistent review process, a set of tools were developed that include a detailed framework capturing all the necessary attributes, a systematically designed online literature review questionnaire and a manual to guide the respondents on completing the questionnaire. In the manual a detailed description on what each attribute means and a clear explanation on how to identify each attribute in a published paper were provided. An additional training event for the reviewers was conducted to make sure that all the participants were comfortable and had a robust understanding of the review process. In addition, to further guarantee that all the attributes were identified correctly, each paper was reviewed by two reviewers; one experienced and one junior. In cases of disagreement on an attribute of the review, two senior reviewers performed a final review of the paper collaboratively so that consensus could be achieved. In this review 123 papers were identified following the screening process and reviewed in order to address a significant gap in the literature. *However, this review has the following limitations*:

1. *Possible missing relevant papers*: Even if this is a representative sample of papers published in both medical and AI journals and conference proceedings, it may not reflect the entire range of the literature regarding BNs in healthcare. We looked for keywords such as *Bayesian Networks*, *probabilistic graphical models*, *medical*, *clinical* to appear in the abstract of each paper. It is possible that a small number of relevant papers were not included because they did not use the selected keywords in their abstract. This is especially true in cases where the actual name of the medical condition is described without mentioning the words *medical* or *clinical*. However, we believe that the large number of selected papers was sufficient for drawing conclusions.
2. *Limited time period*: Because of the plethora of published papers and the time needed for review and data extraction we limited the study to the seven-year period 2012-2018. Although this a safe number of years to be able to draw conclusions, papers published before 2012 were not reviewed. As the review began in early 2019, so we decided not to include papers published in 2019, as we believe that adding a year will have no significant contribution to our findings.
3. *Subjectivity*: Despite the detailed framework, the well-designed online survey and the guiding manual a few of the attributes were subjective and difficult to be identified. For instance, to capture how generic was the overall development process as well as the structure and parameter learning or elicitation process, a subjective 5-scale answer was provided with the labels: *None*, *Implied*, *Some evidence of method*, *High level of description*, and *Repeatable*. For some papers choosing the appropriate state was difficult and it was based on reviewer judgement. For evaluating the clinical usefulness of the BN in practice a set of elements previously described in the literature were examined. Some of these elements were again subjective and many times not clearly stated in the papers. However, to increase the correctness of our findings, each paper was read by two reviewers and a consensus was achieved in case of a disagreement.
4. *Possible missing evidence of adoption*: One of our key findings was the total absence of clinical adoption of any BN in the literature reviewed. However, it is possible that evidence for subsequent translation and adoption of BNs into the real world may well appear in other places, such company reports, media, general commentaries and not necessarily in journal articles. For instance, based on limited marketing materials, there are indications that Babylon is using BNs in their non-public research and application development (Armstrong 2018), including what is claimed as the largest BN in the world (MMC Ventures 2017). As a result, BNs that have been actually used in practice might have been missed. For that reason we tried to infer the adoption of the published BNs in practice either from the overall impression of the paper or based on whether the BN follows all the rules necessary for having a useful model, as a model cannot be applied in practice if it has not been proven to be useful first.

# 8 Recommendations for researchers

In this section, specific recommendations for improving the *reproducibility* of the modelling approach described in published papers as well as the *usefulness* of the BNs in practice are presented.

1. *The clinical benefit of the BN should be explained in depth*: As discussed in the third objective of this scoping review, the clinical benefit of a BN was usually explained in the published papers. However, there were many cases where the benefit was not provided, and therefore the aim of the BN was difficult to understand. The clinical benefit should be stated at the beginning of each paper, preferably supported with relevant literature and statistics that justify the need for a BN capable of performing better than the current practice standard. The effort of developing a BN to be used as a CDSS should be based on the need to address an important clinical problem with a significant potential benefit. Making the extra effort of thinking and describing the benefit from the beginning can help readers better understand the model's aim, and researchers better understand the future impact a model may have. One example of properly representing the benefit of the BN can be seen in the introduction of the work by Ducher et al. (2015).
2. *The structure and the variables of the BN should be provided*: The second objective of this review identified that, in most papers, neither the BN structure nor the variables were clearly described. This makes it difficult to review the BN logic, and hides important knowledge that could potentially be used in other studies. As a result, the complete BN structure should be provided in a paper, and further, the variables of the BN should be fully described. Tables that include the full variable name, a description of the variable and its possible states are extremely useful and can be a significant source of knowledge that improves readers comprehension of the proposed model. An example of how variables should be described in this manner can be found in the papers published by Velikova et al. (2014) and Constantinou et al. (2015).
3. *The parameters of the BN should be provided*: Also, from the second objective of this review we identified that in most papers the BN parameters were not disclosed. Failure to disclose the parameters completely inhibits reproducibility of the claimed results. Additionally, complete NPTs can be a valuable source of knowledge for similar studies. To resolve this issue the marginal prior probabilities and complete NPTs should always be made available to the reader. An illustrative example on how NPTs should be presented can be found in the work of Jochems et al. (2016).
4. *The BN development process should be explained in detail to the point of being repeatable*: We identified in the second objective of this review that the vast majority of published papers failed to fully describe the BN development process. Absence of a clear development process makes the research difficult to follow and impossible to duplicate. For that reason, the overall methodology should be described in a generic way to improve the readers comprehension of the approach used, and their ability to replicate the methodology used. A diagrammatic illustration of the overall methodology can be very informative, such as the one presented in Yet et al. (2014).
5. *When the BN structure is elicited from knowledge, the elicitation process followed should be described in detail to the point of being repeatable*: While reviewing the modelling approach used, we observed that in cases where the BN structure was elicited from knowledge the process was rarely described. To simply say that the BN structure was elicited from experts or literature is not a sufficient description of methodology. The process of extracting variables, as well as the choice of arcs to be used, should be described clearly and open to critique by the reader. An example of a clear and concise description can be found in the works of Akhtar et al. (2014) and Seixas et al. (2014).
6. *When the BN parameters are elicited from knowledge, the elicitation process followed should be described in detail to the point of being repeatable*: Another finding from the second objective of this review was that in many papers where the BN parameters were elicited from knowledge, the process was not described in any detail. As with the previous items discussed above, this limits the reader's ability to understand, critique and replicate the modelling process. Consequently, the process for eliciting the parameters from knowledge should be clearly described both for methodological review, and for repeatability. In cases where the parameters are extracted from experts it is necessary to explain not only how the parameters were elicited, for instance using graphs or anchors, but also the number of experts involved, time and resources consumed by the process, how the different answers among the experts have been considered, the certainty of the elicited subjective parameters and the benefits and limitations of the chosen process. Similarly, when the parameters are elicited from literature a list of the publications that were used should be provided, the process of selecting the necessary parameters should be explained, and the degree of trust for the elicited parameters. An example of a well described parameter elicitation process can be found in Yet et al. (2013) and Constantinou et al. (2015).

7. *When the BN structure and parameters are learned from data, the learning algorithm as well as the dataset used, and the way missing data was treated should be provided*: We identified that insufficient information was usually provided for the characteristics and limitations of the data that was used to develop data-driven BNs. This creates significant gaps in the reader's understanding of the modelling process. When learning the BN structure from data, the dataset used should be described, including presentation of elements such as the dataset origin, size and limitations, as in Lee et al. (2017) and Zador et al. (2016). In addition, the learning algorithm used should be described in detail, such as in the works of Lee et al. (2017) and Luo et al. (2017). A significant element that should never be overlooked is how missing data were identified and classified, for example: not missing at random or missing completely at random, and so on. How this missing data was treated, i.e. whether missing values were computed or ignored, should also be described. A clear description of the learning process makes it easier for readers to follow and potentially replicate the process. A good example of a description of data pre-processing and how to handle missing data can be found in the paper published by Sesen et al. (2013).
8. *The credibility of a data-driven BN structure should be explained*: From the third objective of this review we learned that the credibility of data-driven BNs was difficult to assess as it was usually underexplored by authors. When the BN structure is learned from data, it represents the relationships that are available in the dataset, which are not always in agreement with the true causal mechanisms of a given disease. For that reason, a justification of the model's credibility is needed to ensure that the BN has a clear logic and evaluates the answer it was intended to compute. Representing true causal/ influential relationships is particularly important when the BN is intended for interventional or counterfactual reasoning. An example of credible data-driven BN structures can be found in the works of Chao et al. (2017) and Kaewprag et al. (2017).
9. *A validation of the model's accuracy should be provided*: As observed in the third objective of this review, most authors provided some claim of their model's accuracy. However, their approach to assessing and validating accuracy was not always described with sufficient clarity to be easily understood and evaluated by the reader. In this review we found several ways to validate a BN. A simple approach for evaluating the model's reasoning was through the use of specific cases or scenarios. Although useful, this is not sufficient to fully validate a model's accuracy. The minimum validation process needed is internal validation, where the predictive performance of the model is assessed using resampling techniques such as bootstrapping or cross-validation. All steps followed in validation of the BN should be explained in sufficient detail to enable repeatability. In cases where low accuracy is observed, justification and potential corrections should be offered. In cases where internal validation has not been performed, this should be openly disclosed and suggested as future work, thus indicating the need for a validation study. A well-described validation and review process can be found in Yet et al. (2014).
10. *An external validation to evaluate the model's generalizability should be considered*: An important finding of the third objective of this review was a lack of generalizability for most of the described models. An accurate model may not be useful in clinical practice if it is not proven to work consistently on different populations. Thus, an external validation is necessary for validating the model's performance on different datasets. Providing an external validation significantly strengthens the performance and utility of a BN. In cases where an external validation has not been performed due to data or time restrictions, it should be openly disclosed and suggested as future work as we see in the work of Park et al. (2018). An example of external validation can be found in the paper published by Luo et al. (2018).
11. *The actual use of the model in practice and its impact on healthcare should be considered*: The third objective of our review identified that impact analysis was usually neglected, especially in research papers where the sole focus was on development of the BN. However, in order to bridge the gap between having accurate models and delivering impact in practice, more focus should be given to identifying issues and approaches to help clinicians adopt the decision support model. In cases where a small pilot study to investigate the potential impact of the models, such as in the work of Mcheick et al. (2017), or a properly randomized control trial have not been performed, this should also be openly disclosed and identified as potential future work. Acknowledging the need for an impact study and how the model could be applied in practice to help clinical decision making provides the reader with a clear understanding of the intended use and the potential benefits that may result from use of the proposed model. A representative example acknowledging an impact analysis as a crucial future step is available in Velikova et al. (2014).

# 9 Future Directions

This scoping review revealed numerous research areas that have been neglected, and that should receive further attention in the future. The following are some of the key future research areas that emerge from this study:

1. *Multi-disciplinary expert team involvement in BN development:* When developing medical BNs intended for use as CDSS there is a need for collaboration between clinicians and decision scientists. Clinical expertise is important to ensure that the clinical benefit and adoption of the model are considered, while decision scientists, computer scientists or statisticians are necessary to develop an accurate BN model. Methodologies and tools to support multidisciplinary expert involvement processes for BN development would be useful to resolving this need.
2. *Using BNs to answer hypothetical questions:* BNs should be used to perform more than just reasoning from evidence. They can also be used to answer *hypothetical questions.* More research should focus on modeling and validating medical BNs for performing *interventional* and *counterfactual reasoning.*
3. *Temporal BN models:* Greater effort is needed to identify better methods for developing temporal BNs that capture the evolving process of clinical decision making.
4. *Clinical datasets for use in harnessing ML for developing BNs:* Despite supposedly being in the era of big data and machine learning, many papers in this review identified a lack of good quality clinical datasets as a barrier to developing accurate BNs. Solving the issues of data integrity, integration and interoperability, especially in electronic health records and routine clinical datasets, will be key to ensuring that health data can be used to render credible and accurate predictions and decisions for use in the care of individual patients and their families.
5. *Mixed methodologies for developing BNs:* More work is needed to develop BNs using mixed approaches that are based on both data and knowledge.
6. *Tools and methods for eliciting expert knowledge for BN development:* Expert elicitation remains largely *ad hoc* and research into more methodological approaches is needed in order to standardize these processes.
7. *Adoption of medical BNs in clinical practice:* While there has been significant research interest in developing accurate medical BNs, their adoption in clinical practice is less obvious. More attention should be paid to the process of translating an accurate BN to a useful and usable CDSS that has potential for benefit when applied in clinical practice.

# 10 Summary and Conclusions

BNs can be successfully applied to model complex medical problems requiring reasoning under uncertainty. An immense interest is demonstrated by the volume of literature proposing medical BNs and the research effort expended globally in developing BNs to support clinical decision making. This scoping review, drawing on 123 papers that propose BNs in healthcare, evaluated the decision support, development and usefulness of BN models proposed for use in clinical practice. We found that the majority of published BNs focus on supporting simple decisions where reasoning from evidence is required, with authors neglecting more complex reasoning approaches that BNs may be applied to, and the dynamic nature of clinical decision making. There is a preference for data-driven BNs even though medical datasets usually lack integrity, presenting with poor quality and incomplete. There is also an important lack of repeatable modelling approaches and a significant gap between development of a claimed accurate model, and demonstration of a clinically useful model capable of delivering actual impact on healthcare. Consequently, we have proposed several future research directions and recommendations for improving reproducibility of modelling approaches described in the literature, and for improving the overall usefulness of medical BNs. With these future directions we hope the quality and repeatability of published papers on BNs in healthcare will improve and that all researchers in this domain can work towards closing the chasm between research interest and clinical adoption.

## ACKNOWLEDGMENTS

EK, SM, MRN, AF and NF acknowledge support from the Engineering and Physical Sciences Research Council (EPSRC) under project EP/P009964/1: PAMBAYESIAN: Patient Managed decision-support using Bayes Networks. KD acknowledges financial support from the School of Fundamental Sciences, Massey University, New Zealand, for his study sabbatical and visits with the PamBayesian team.